\documentclass{article} 
\usepackage[final]{colm2025_conference}

\usepackage{lineno}
\usepackage[utf8]{inputenc} 
\usepackage[T1]{fontenc}    
\usepackage{hyperref}       
\usepackage{url}            
\usepackage{booktabs}       
\usepackage{amsfonts}       
\usepackage{nicefrac}       
\usepackage{microtype}      
\usepackage{xcolor}         
\usepackage{enumitem}
\usepackage{natbib}
\usepackage{wrapfig}
\usepackage{float}    
\usepackage{subcaption} 
\newcommand{\wco}{\text{word-concept}\xspace}
\newcommand{\cco}{\text{context-concept}\xspace}

\newcommand{\Wbar}{\overline{\W}}
\newcommand{\Hbbar}{\overline{\Hb}}
\newcommand{\Sigmabbar}{\overline{\Sigmab}}

\newcommand{\ubd}{\ub^{\rm{d}}}
\newcommand{\vbd}{\vb^{\rm{d}}}

\newcommand{\Winf}{\W_\infty}
\newcommand{\Hbinf}{\Hb_\infty}

\usepackage[mathscr]{euscript}

\newcommand{\Hmm}{\Hb^{\rm{mm}}}

\newcommand{\pbh}{\hat{\pb}}

\newcommand{\Wmm}{\W^{\rm{mm}}}

\newcommand{\Lbmm}{\Lb^{\rm{mm}}}

\newcommand{\smat}{\Sb}

\newcommand{\smatbar}{\widetilde{\smat}}
\newcommand{\Sbar}{\widetilde{\smat}}
\newcommand{\sbar}{\widetilde{\s}}

\newcommand{\Lb}{\vct{L}}

\usepackage{hyperref,graphicx,amsmath,amsfonts,amssymb,bm,url,breakurl,epsfig,epsf,color,fmtcount,semtrans,caption,multirow,comment, boldline, amsthm}
\usepackage{subcaption}
\usepackage{bm}
\usepackage{enumitem}
\usepackage{amssymb}

\setlist[itemize]{leftmargin=5mm}

\usepackage{hyperref}
\usepackage[utf8]{inputenc} 
\usepackage[T1]{fontenc}    
\usepackage{url}            
\usepackage{booktabs}       
\usepackage{nicefrac}       
\usepackage{microtype}      
\usepackage{dsfont}
\usepackage{xspace}

\newcommand{\vb}{{\vct{v}}}

\newcommand{\Vb}{{\mtx{V}}}

\newcommand{\Pbb}{{\mtx{P}}}

\newcommand{\vct}[1]{\bm{#1}}
\newcommand{\mtx}[1]{\bm{#1}}

\usepackage{mathtools}

\usepackage{titlesec}

\usepackage{tikz}
\usepackage{pgfplots}
\usetikzlibrary{pgfplots.groupplots}

\usepackage{movie15}

\usepackage{caption}
\usepackage[bottom,hang,flushmargin]{footmisc} 

\newcommand{\tsn}[1]{{\left\vert\kern-0.25ex\left\vert\kern-0.25ex\left\vert #1 
    \right\vert\kern-0.25ex\right\vert\kern-0.25ex\right\vert}}

\definecolor{darkred}{RGB}{150,0,0}
\definecolor{darkgreen}{RGB}{0,150,0}
\definecolor{darkblue}{RGB}{0,0,200}
\hypersetup{colorlinks=true, linkcolor=darkred, citecolor=darkgreen, urlcolor=darkblue}

\newtheorem{theorem}{Theorem}

\newtheorem{definition}{Definition}

\newcommand{\Rb}{\mathbf{R}}

\newcommand{\diag}[1]{\operatorname{diag}(#1)}

\newcommand{\appropto}{\mathrel{\vcenter{
  \offinterlineskip\halign{\hfil$##$\cr
    \propto\cr\noalign{\kern2pt}\sim\cr\noalign{\kern-2pt}}}}}
    
\newcommand{\cut}[1]{\textcolor{red}{}}
\newcommand{\W}{{\vct{W}}}
\newcommand{\Ab}{{\vct{A}}}

\newcommand{\sign}[1]{\texttt{sign}(#1)}

\newcommand{\Ub}{\vct{U}}

\newcommand{\Hb}{{\vct{H}}}
\newcommand{\Sb}{\vct{S}}

\newcommand{\A}{\vct{A}}

\newcommand{\Pb}{\mathds{P}}

\newcommand{\pb}{\vct{p}}

\newcommand{\ub}{\vct{u}}

\newcommand{\eb}{\vct{e}}

\newcommand{\s}{\vct{s}}
\newcommand{\z}{\vct{z}}

\newcommand{\ab}{\vct{a}}

\newcommand{\hb}{\vct{h}}

\newcommand{\Vc}{\mathcal{V}}

\newcommand{\Kc}{\mathcal{K}}

\newcommand{\Lc}{\mathcal{L}}

\newcommand{\beq}{\begin{equation}}
\newcommand{\eeq}{\end{equation}}
\newcommand{\bea}{\begin{align}}
\newcommand{\eea}{\end{align}}

\newcommand{\R}{\mathbb{R}}

\newcommand{\nn}{\notag}

  \newcommand{\Sigmab}{\boldsymbol\Sigma}
    
  \newcommand{\la}{{\lambda}}                     

\DeclarePairedDelimiterX{\inp}[2]{\langle}{\rangle}{#1, #2}

\newcommand{\Id}{\mathds{I}}
\newcommand{\ones}{\mathds{1}}

\definecolor{darkblue}{rgb}{0, 0, 0.5}
\hypersetup{colorlinks=true, citecolor=darkblue, linkcolor=darkblue, urlcolor=darkblue}

\title{Geometry of Semantics in Next-Token Prediction: \\
How Optimization Implicitly Organizes \\Linguistic Representations}

\author{Yize Zhao, Christos Thrampoulidis\\
Department of Electrical and Computer Engineering\\
University of British Columbia\\
Vancouver, BC, Canada \\
\texttt{\{zhaoyize, cthrampo\}@ece.ubc.ca} \\
}

\date{}

\begin{document}
\maketitle
\begin{abstract}
We investigate how next-token prediction (NTP) optimization leads language models to extract and organize semantic structure from text. Our analysis, based on a tractable mathematical model and controlled synthetic data, reveals that NTP implicitly guides models to factor a centered support matrix encoding context-to-next-token co-occurrence patterns via singular value decomposition (SVD). While models never explicitly construct this matrix, learned word and context embeddings converge to its SVD factors, with singular vectors encoding latent semantic concepts through their sign patterns. We demonstrate that concepts corresponding to larger singular values are learned earlier during training, yielding a natural semantic hierarchy where broad categories emerge before fine-grained ones. This insight motivates orthant-based clustering, a method that combines concept signs to identify interpretable semantic categories. We validate our findings on synthetic datasets and pretrained language models, recovering diverse semantic structures such as grammatical categories, named entity types, and topical distinctions (medical, entertainment). Our work bridges classical distributional semantics and neural collapse geometry, characterizing how gradient-based optimization implicitly determines both the matrix representation and factorization method that encode semantic structure.
\end{abstract}

\section{Introduction}

Models trained with next-token prediction (NTP) show a remarkable ability to capture and represent linguistic meaning. \emph{How does this semantic ability emerge during training?}
A primary challenge in addressing this question lies in quantifying ``meaning''. We therefore focus specifically 
on understanding the semantic organization of learned representations and the mechanisms by which it arises.
Our goal is fundamentally different from the substantial recent interest in post-hoc evaluation of interpretable semantics in Large Language Models (LLMs). Techniques like dictionary learning and sparse autoencoders for model steering \cite{cunningham2023sparse, bricken2023monosemanticity, turner2023activation, li2024inference}—tracing back to similar work with static word embeddings for word2vec architectures \cite{murphy2012learning, faruqui2015sparse, arora2018linear, zhang2019word}—aim to extract interpretable features from trained models. In contrast, we seek to understand the specific mechanisms by which gradient-descent-based training, in conjunction with the structure of textual input data, guides a language model to organize its representations into semantically interpretable structures.

Our starting point is the recent work by \citet{zhao2024implicit}, who  characterized the geometry of the model's \emph{explicit} textual input representations, i.e., $d$-dimensional word/context embeddings. At convergence, this geometry is precisely characterized by the factorization of a specific \emph{data-sparsity matrix} $\smatbar$, which encodes co-occurrence patterns in the training corpus: its entries indicate whether a particular word follows a given context.

This characterization, however, leaves open the question: \textbf{How do these textual input patterns impact the geometry of \emph{implicit} linguistic information}, such as grammatical or semantic categories, \textbf{through the optimization of the NTP objective?}
Specifically, we ask three subquestions:

Since semantic information is not explicitly labeled in the data, 

\noindent\textbf{Q1.} \emph{what latent signals implicitly guide NTP optimization to specific structures?}
 
\textbf{Q2.} \emph{How are these structures related to human-interpretable semantics?}
 
\textbf{Q3.} \emph{How fast do these semantic structures emerge during training?}\\ 

\begin{figure*}[t!]
    \centering
    \includegraphics[width=1.0\linewidth]{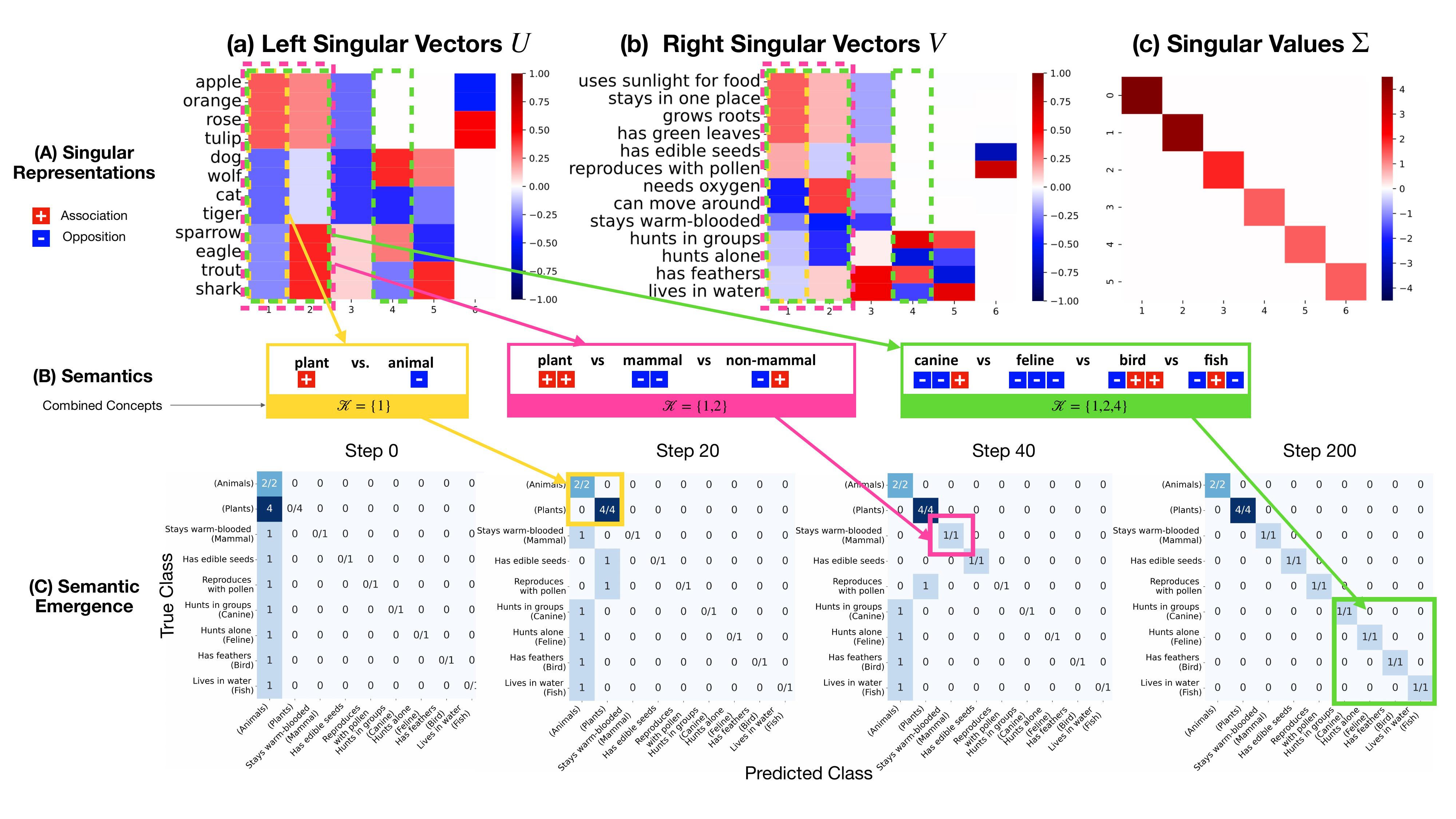}
        \captionsetup{width=\textwidth}
\caption{\textbf{Illustration of C1-3.} Details in Sec. \ref{sec:concept geometry}. \textbf{(A)} SVD of the data-sparsity  matrix $\smatbar$ (Eq. \eqref{eq:smatbar}), derived from the training data in  Fig. \ref{fig:motivation}(C). Columns of $\Ub$ and $\Vb$, serve as word and context {concept basis} vectors, respectively: Their entries encode an association score with sign indicating the direction (association/opposition) between word/contexts and concepts. The singular values indicate the strength of each concept. 
 \textbf{(B)}  Combining  a few top concepts according to the specific sign patterns of concept basis, which we call  orthant-based clustering, reveals human-interpretable semantics. \textbf{(C)} Gradient-descent-based NTP optimization leads to concepts corresponding to larger singular values being learned first. This corresponds to the emergence of coarse semantics before fine-grained ones, which we empirically validate with visualizations of the trained model's output confusion matrices.
}
    \label{fig:concept_figures}
\end{figure*}
Towards addressing these questions, we adopt a bottom-up, theory-guided approach. We build our insights by analyzing the optimization of an abstract mathematical model, which isolates the essential components of NTP optimization in a tractable framework (Sec. \ref{sec:background_main}). To validate and further shape these insights, we carefully construct synthetic data settings and formally verify our findings, which can otherwise be difficult to precisely quantify in complex, real-world datasets (Sec. \ref{sec:concept geometry}). Finally, we demonstrate that the insights gained extend qualitatively to  real data and models (Sec. \ref{sec:experiments}).
Concretely, our contributions are:



\noindent
\textbf{C1. Interpreting latent signals (Sec.~\ref{sec:svd}):}  
We show that the singular vectors of $\tilde{\mathbf{S}}$ encode latent concepts: the sign of each entry indicates whether a word or context supports or opposes a concept, and the magnitude reflects the strength of association. This equips the abstract geometry in \cite{zhao2024implicit} with semantic structure.
Fig.~\ref{fig:concept_figures}A  visualizes semantics  aligning with concept signs.

\noindent\textbf{C2. Establishing human-interpretable semantic structure (Sec.~\ref{sec:orthant}):}  
Building on C1, we introduce an orthant-based clustering method that identifies interpretable semantic categories by combining a few top concepts according to their sign patterns.
We empirically demonstrate that this method recovers a wide array of interpretable linguistic structures—such as part of speech, verb tense, and named entity types—on both synthetic data and real datasets like TinyStories and WikiText-2, where the data-sparsity matrix is explicitly computable.
Furthermore, we show in GPT-2, BERT, and Qwen that increasing the number of singular components systematically reveals finer-grained semantics. Fig.~\ref{fig:concept_figures}B visualizes  a hierarchy of interpretable semantics via orthant-based clustering.  

\noindent\textbf{C3. Analyzing semantic emergence (Sec.~\ref{sec:rate}):}  
Beyond  convergence, we analyze semantic emergence during training. Theoretically, we link NTP-UFM gradient descent dynamics to the framework of \citet{saxe2013exact}, showing that top concepts, those corresponding to larger singular values, are learned earlier. Fig.~\ref{fig:concept_figures}C visualizes this phenomenon: broad semantics corresponding to top concepts are formed early during training. 


These contributions bridge classical distributional semantics (e.g., \cite{ding2006nonnegative,levy_semantics}) and recent  neural collapse (NC) geometries \cite{NC,zhao2024implicit}. By employing tractable abstractions from the latter, we provide theoretical grounding for the former: we show that gradient-based optimization implicitly determines both the matrix representation (data-sparsity matrix) and factorization method (SVD) for encoding semantics, rather than requiring hand-engineered choices (Sec. \ref{sec:related_work_main}). Conversely, we extend the scope of NC geometries beyond simple classification settings to characterize the underlying hierarchical semantic representations.

\noindent\textbf{Notations.} For any integer $k$, $[k]:= \{1,\ldots,k\}$. Matrices, vectors, and scalars are denoted by $\A$, $\ab$, and $a$, respectively. For matrix $\A$, $\A[i,j]$ is its $(i,j)$-th entry, and for vector $\ab$, $\ab[i]$ is its $i$-th entry. $\ones$ is the all-ones vector and $\Id$ is the identity matrix.

\section{Background: Geometry of Word and Context Representations}
\label{sec:background_main}

The NTP objective minimizes the cross-entropy (CE) loss between the model's output distribution given a context—a sequence $\z$ of tokens from vocabulary of size $V$—and the one-hot representation of the next-word/token. Since multiple tokens can follow a given \emph{distinct} context $\z_j$ in a dataset, this is equivalent to minimizing  CE  between the model's output distribution and an associated \emph{soft-label vector} $\pbh_j$ representing the conditional frequency distribution of next tokens for that specific context. Because not all vocabulary tokens are valid next-tokens for a given context (even when ignoring the finite size of the data), the soft-label vectors $\pbh_j$ are \emph{sparse}. For a dataset with $m$ distinct contexts, define the   \textbf{support matrix} $\smat\in\{0,1\}^{V\times m}$ of the soft-label matrix $\Pbb=[\pbh_1,\ldots,\pbh_m]\in\R^{V\times m}$, such that  $\smat[z,j]=1\,\Leftrightarrow\,\Pbb[z,j]>0$. 

Building on this formulation, \citet{zhao2024implicit} investigated how the soft-label structure shapes the learned geometry when optimizing the NTP objective with gradient descent. Specifically, they studied word embeddings (the rows of the $V\times d$ decoding matrix $\W$) and context embeddings (the deep $d$-dimensional representations $\{h(\z_j)\}_{j\in[m]}$ of distinct contexts) of a model at convergence.

For tractability, they relied on two modeling simplifications: (1) using the limit of \emph{vanishing ridge regularization} as a proxy for the convergence limit of gradient-descent iterations, and (2) assuming a sufficiently expressive neural network that optimizes context embeddings \emph{freely}, instead of abiding by an architecture-specific parameterization. 
The first simplification is standard in the implicit-bias optimization literature \cite{ji2020gradient,wei2019regularization,tarzanagh2023transformers}, while the second has been used for language models by \citet{yang2017breaking} and recently popularized in the neural collapse literature \citet{mixon2022neural,wojtowytsch2021emergence,fang2021exploring}.

Taken together, these abstractions yield a tractable model for NTP optimization, which we refer to as the unconstrained features model (UFM) for NTP training, or \textbf{NTP-UFM} in short:\footnote{Note the structural similarity of this log-bilinear objective to word2vec architectures \cite{word2vec_1,glove}. However, NTP-UFM  serves as a tractable analytical abstraction rather than a direct practical architecture.}
\begin{align}\label{eq:ufm}
    \min_{\W,\Hb}\Lc(\W\Hb;\Pbb)+{\la}\|\W\|^2 + {\la}\|\Hb\|^2\,.
\end{align}
NTP-UFM jointly optimizes the matrices of \textbf{word embeddings} $\W\in\R^{V\times d}$ and \textbf{context embeddings} $\Hb:=[\hb_1,\ldots,\hb_m]\in\R^{d\times m}$. Here, $\hb_j$ represents $h(\z_j)$ and is freely optimized for each context; $\la$ is the regularization parameter; and $\Lc$ is the CE loss acting on the logit matrix $\Lb=\W\Hb$,  parameterized by the soft-label matrix $\Pbb$.
Analyzing NTP-UFM, for large embedding dimensions $d\geq V$, \footnote{
This does not restrict the number of contexts ($m$, potentially $\gg d$), enabling the study of rich semantic catgories (Sec.~\ref{sec:concept geometry}). Also, qualitative insights extend to real models (Sec.~\ref{sec:exp_findings}).
} and $\lambda\rightarrow0$, \citet{zhao2024implicit} show that:

     \noindent\emph{1. Logits:} The logit matrix $\Lb$ grows unboundedly during training. After normalization, it converges in direction to a matrix $\Lbmm$ that can be explicitly computed given the data support matrix $\Sb$.

    \noindent \emph{2. Word/Context Embeddings:} Like logits, word and context embeddings grow unboundedly in magnitude, but converge in direction to $ \Ub\sqrt{\Sigmab}\Rb$ and  $ \Rb^\top\sqrt{\Sigmab}\Vb^\top$, respectively. Here, $\Ub\Sigmab\Vb^\top$ is the SVD of $\Lbmm$ and $\Rb$ is a partial orthogonal matrix.
    
    \noindent \emph{3. Data-sparsity matrix as proxy:} An accurate proxy for $\Lbmm$ that is very easy to compute is the centered  support matrix
    \begin{align}\label{eq:smatbar}
        &\smatbar:=(\Id_V-\nicefrac{1}{V}\cdot\ones_V\ones_V^\top)\,\smat\, \\ &\smatbar[z,j]=\begin{cases}
        1-\frac{1}{V}\sum_{z\in[V]}\smat[z,j] & \text{if $\smat[z,j]=1$}\,
        \\
        -\frac{1}{V}\sum_{z\in[V]}\smat[z,j] & \text{if $\smat[z,j]=0$}\,\nn
    \end{cases}\,,
    \end{align}
    which we  refer to  as the \textbf{data-sparsity matrix}. Note that unlike the support matrix $\smat$, the entries of $\smatbar$ are not binary due to the centering operation.

Taken together, \citet{zhao2024implicit} establishes how textual input structure maps through NTP optimization to word and context representation geometry. Textual input structure is summarized through the SVD of the data-sparsity matrix, where $\Ub\in\R^{V\times r}$, $\Vb\in\R^{m\times r}$ with $\Ub^\top\Ub=\Vb^\top\Vb=\Id_r$, singular values $\Sigmab=\diag{\sigma_1,\ldots,\sigma_r}$ are ordered as $\sigma_1 \geq \ldots \geq \sigma_r >0$, and rank $r\leq V-1$:
\begin{align}\label{eq:smatbar svd}
\Sbar=\Ub\Sigmab\Vb^\top.
\end{align}
The geometry of learned representations is characterized by word and context embeddings that converge in the direction to $\Ub\sqrt{\Sigmab}\Rb$ and $\Rb^\top\sqrt{\Sigmab}\Vb^\top$, respectively, with $\Rb$ a partial orthogonal matrix.

\begin{figure*}[t]
    \centering
    \begin{tikzpicture}
        \def\xgap{4.8}
        \node[inner sep=0pt] at (0, 0) {
            \includegraphics[width=0.3\textwidth]{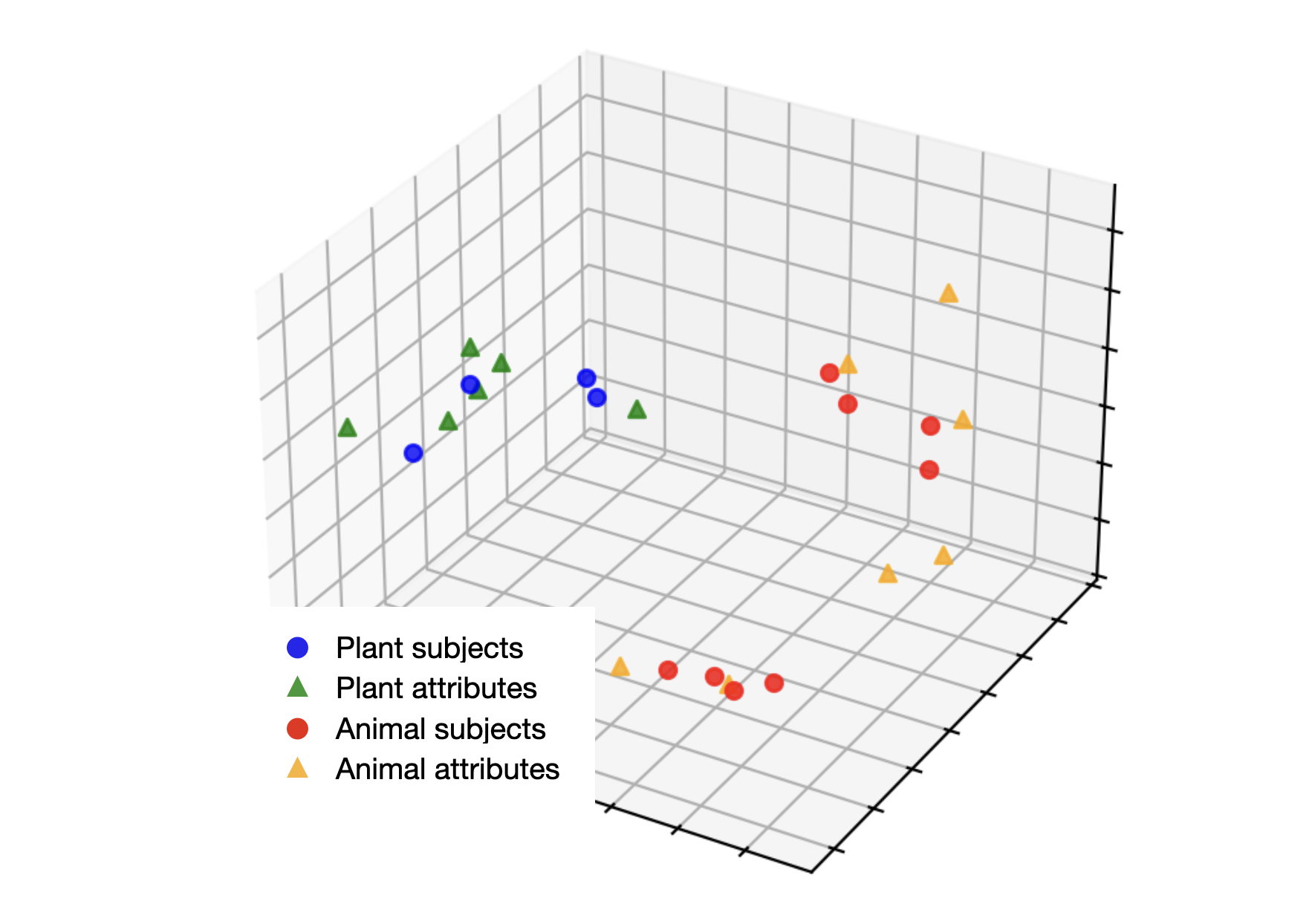}
        };
        \node at (0, 2.2) {\textbf{(A) Embedding geometry}};

        \node[inner sep=0pt] at (\xgap, 0) {
            \includegraphics[width=0.34\textwidth]{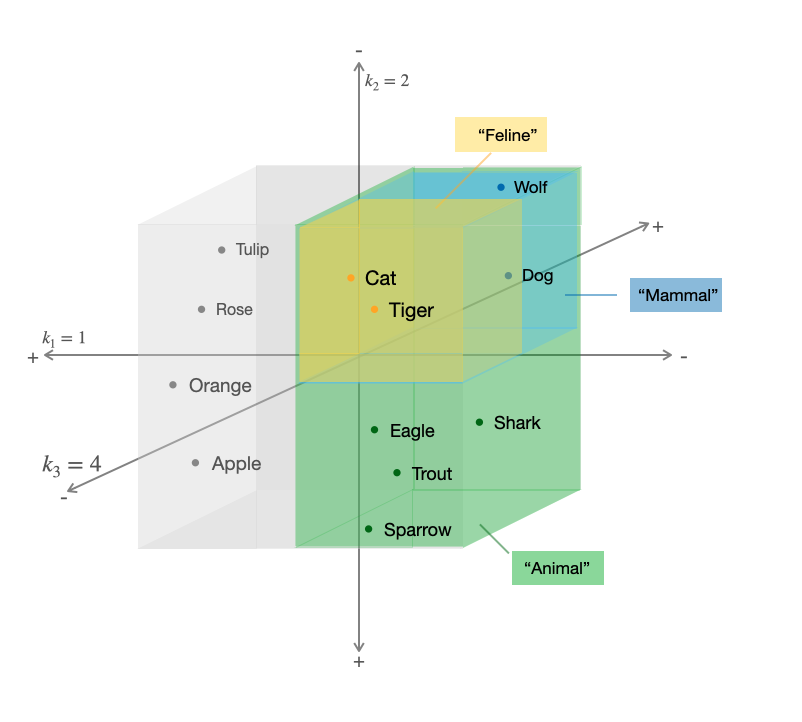}
        };
        \node at (\xgap, 2.2) {\textbf{(B) Orthant-based clustering}};

        \node[inner sep=0pt] at (2*\xgap, 0) {
            \includegraphics[width=0.38\textwidth]{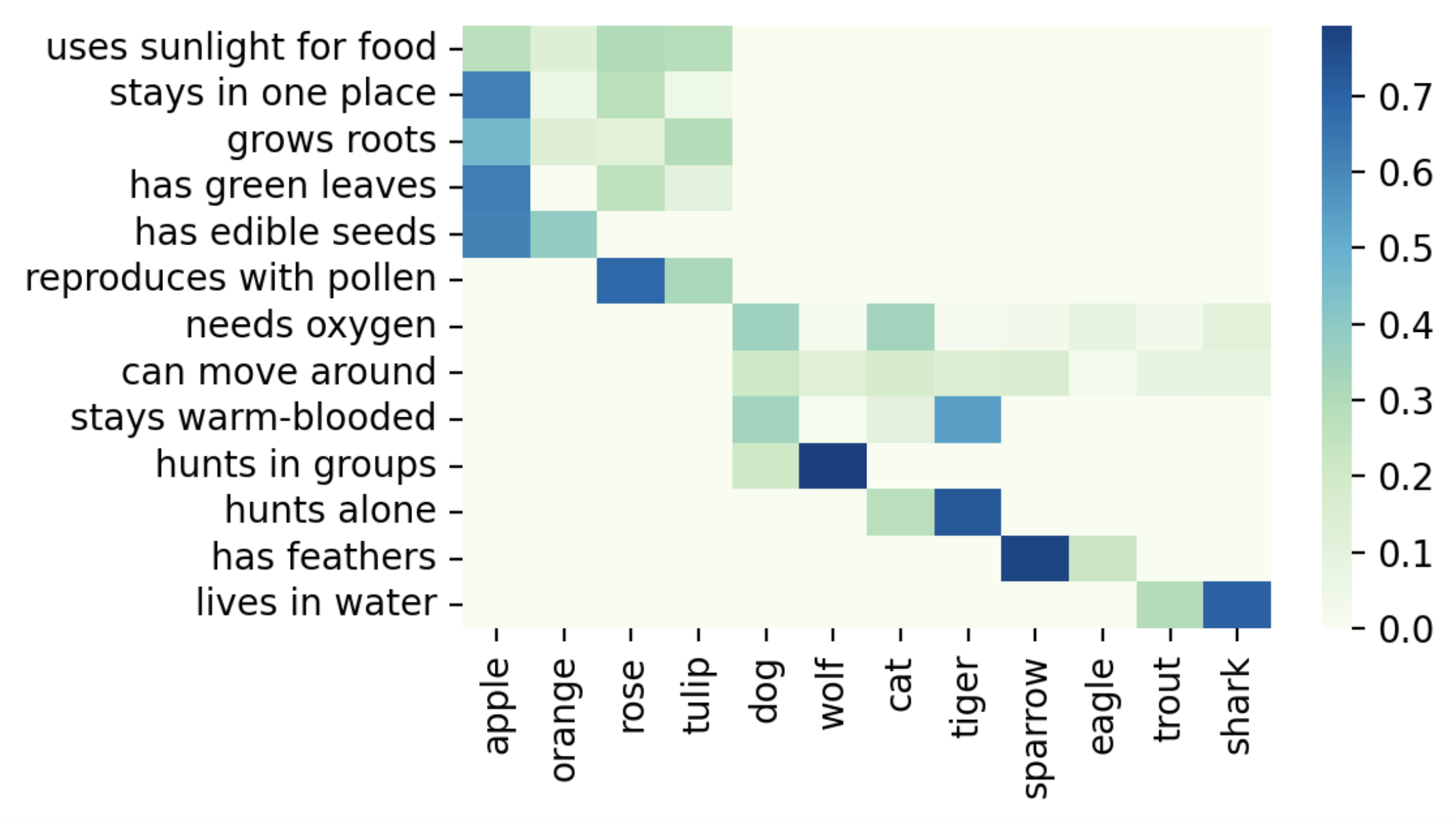}
        };
        \node at (2*\xgap, 2.2) {\textbf{(C) Training data}};
    \end{tikzpicture}
    \captionsetup{width=\textwidth}
\caption{ \textbf{Optimizing NTP on (context,next-word) input pairs yields a specific geometry of representation of those inputs; orthant-based clustering reveals how this geometry is organized on the basis of interpretable semantic categories.}
\textbf{(A)} Empirical 3D PCA visualization of learned word(dots)/context(triangular marks) embeddings from two-layer transformer trained on the dataset in (C) reveals a clear clustering by semantic category (plants in green/blue, animals in red/orange), despite no explicit semantic signal at training.
\textbf{(B)} We show (Sec. \ref{sec:orthant}) that semantic categories are organized into a hierarchy of sub-orthants defined by concepts combined with respect to their sign patters. For example, Felines occupy a 2D sub-orthant (combining the second and fourth concepts), while Mammals and Animals occupy progressively larger sub-orthants.
\textbf{(C)} The \emph{sparse} training data matrix (rows = attributes, columns = subjects) follows the template: “The organism that [attribute] is [subject].”
}
\label{fig:motivation}
\end{figure*}

\section{Geometry of Semantics}\label{sec:concept geometry}
The previous section establishes that word and context embeddings converge to the SVD factors of the data-sparsity matrix $\smatbar$. But: (i) \emph{What latent signals implicitly guide optimization to produce semantic structures within this geometry?} (ii) \emph{How are these related to human-interpretable semantics?} (iii) \emph{How fast do they emerge during training?}

To motivate our analysis, consider Fig.~\ref{fig:motivation}A visualizing word/context embeddings of a transformer trained on (sequence,next-token) pairs from the sparse soft-label matrix $\Pbb$ shown in Fig.~\ref{fig:motivation}C. Despite no explicit semantic supervision during training, clear distinctions emerge. For example, animal-related words and contexts cluster on the left, while plant-related ones cluster on the right. What drives such an implicit semantic structure?

\subsection{Identifying Concepts as Latent Signals} \label{sec:svd}

Our key insight is to examine the structure of the SVD factors of the data-sparsity matrix $\smatbar$. Recall  word embeddings take the form $\W = \Ub \sqrt{\Sigmab} \Rb$, where $\Ub$ encodes the row-space of $\smatbar$. Since rows of $\smatbar$ (and, thus, of $\Ub$) correspond to different words, their inner product  captures word similarity: words appearing in similar contexts have high similarity. However, semantic relationships extend beyond simple pairwise similarity or synonymy.

To reveal such semantic structure, we instead focus on the \emph{columns} of $\Ub$, which correspond to $r$ latent features in the representation space. We term each SVD dimension $k\in[r]$ a \emph{concept}. Adopting terminology from \citet{saxe2019mathematical}, we call the columns $\mathbf{u}_k \in \mathbb{R}^V$ and $\mathbf{v}_k \in \mathbb{R}^m$ of matrices $\mathbf{U}$ and $\mathbf{V}$ the "word analyzer vectors" and "context analyzer vectors," respectively. These vectors represent the alignment of words and contexts with each concept: for a word $z$ and concept $k$, the sign of $\mathbf{u}_k[z]$ indicates \emph{association} ($\mathbf{u}_k[z]>0$) or \emph{opposition} ($\mathbf{u}_k[z]<0$) to the concept, while its magnitude quantifies the strength of this relationship. Words for which $|\ub_k[z]|$ is small are \emph{neutral} to this concept. The same interpretation applies to context components $\mathbf{v}_k[j]$. 
We can explicitly represent concepts in the embedding space by defining $d$-dimensional concept representations $\ubd_k=\W^\top\ub_k$ and $\vbd_k=\Hb\vb_k$ for $k\in[r]$ as projections onto the word and context spaces, respectively.
These can also be viewed as weighted averages of word or context embeddings (similar to classical semantic axes) where the entries of $\ub_k$ (or $\vb_k$) reflect each word's (or context's) relevance to the concept.

For an initial confirmation of this interpretation of SVD factors, we revisit the  dataset with syntax "The organism that [attribute] is [subject]",\footnote{For visual clarity, this design isolates semantic concept extraction by minimizing grammatical influences.} shown in Fig.~\ref{fig:motivation}C. Fig.~\ref{fig:concept_figures}A shows the columns of $\Ub$ and $\Vb$, that is, the word and context analyzer vectors. Inspecting them reveals how they can capture semantic information through their sign patterns: For example, the first column/dimension is such that plant-related words and contexts have positive entries, while animal-related ones have negative entries, indicating that this concept represents the semantic distinction between plants and animals.

However, the figure also reveals that \emph{not all} individual concepts correspond to human-interpretable categories. Instead, we posit that interpretable semantics emerge from specific combinations of several concepts. We formalize this in the next section and demonstrate in Sec.~\ref{sec:experiments} that it yields interpretable semantics on larger-scale datasets.

\subsection{Interpretable Semantics via Orthant-based Clustering
}\label{sec:orthant}
We  extend the sign-based interpretation of individual concepts to combinations of multiple concepts. Our key claim is that combining a few top concepts according to their sign patterns, a method we call \emph{orthant-based clustering}, yields interpretable semantic categories. 

For a selected set $\mathcal{K} = \{k_1, \ldots, k_p\}$ of $p\leq r$ concepts, there are $2^p$ possible sign configurations $C = [c_{k_1}, \ldots, c_{k_p}]\in\{\pm1\}^p$, where $c_{k_i}\in\{\pm1\}$ indicates whether dimension $k_i$ contributes positively or negatively to a semantic category. Each configuration defines a potential semantic category with member words and their degrees of association with that category, formalized as follows:
\begin{definition}\label{defn:typicality}
For $C=[c_{k_1}, \ldots, c_{k_p}]$ where each $c_{k_i}=\pm1$ for concept dimensions $k_i\in[r]$: 
\\
$\bullet$ A word $z$ is a member of $C$ if and only if $\sign{\ub_{k_i}[z]}=c_{k_i}$ for all $k_i, i\in[p]$.
\\
$\bullet$ The typicality of a member $z$ of $C$ is defined as:
$
{\rm{Typicality}}(z; C) = \sum\nolimits_{i\in[p]}\big|\ub_{k_i}[z]\big|\,.
$
\\
Analogous definitions apply to contexts.
\end{definition}
Geometrically, configuration $C$ corresponds to an orthant in the $p$-dimensional subspace spanned by  selected concepts. Membership indicates which word embeddings lie within this orthant, while typicality measures the $\ell_1$-norm of a word's projection onto the selected concept dimensions.

As we demonstrate in Sec.~\ref{sec:experiments}, forming orthants with a small number of top concepts yields interpretable semantic categories. Fig.~\ref{fig:motivation}B provides an initial illustration with selected concept dimensions $\{k_1, k_2, k_3\} = \{1,2,4\}$ (i.e., the 1st, 2nd, and 4th columns of $\Ub$ from Fig.~\ref{fig:concept_figures}A(a)).
Words shown with the same color within each orthant are example members according to Defn.~\ref{defn:typicality}. Colored blocks represent different sign configurations: Green ($p=1$) with $c_{k_1}=-1$ captures the "animal" semantic; Blue ($p=2$) with $[c_{k_1},c_{k_2}]=[-1,-1]$ refines this to "mammal"; Yellow ($p=3$) with $[c_{k_1},c_{k_2},c_{k_3}]=[-1,-1,-1]$ further specifies "feline". This nested structure illustrates how increasing the number of combined dimensions yields progressively finer-grained categories:  "feline"$\subset$"mammal"$\subset$ "animal".

This reveals a natural hierarchical structure in orthant-based clustering, where the number of concept dimensions determines semantic granularity: broad categories (e.g., verbs) require fewer dimensions, while specific categories (e.g., past-tense verbs) need more dimensions. See also Sec.~\ref{sec:experiments}.

\subsection{Semantic Emergence}
\label{sec:rate}
{We now show that the semantics' hierarchical structure  is reflected in their emergence during training: \emph{broad categories are learned earlier than fine-grained ones.}  We formalize this by tracking the rate at which concepts are acquired throughout training.  We find that \emph{concepts aligned with dominant singular directions (those corresponding to larger singular values) are learned first.}} Our analytic NTP-UFM model quantifies this claim as follows.

\begin{theorem}\label{thm:main}
    Consider gradient-descent minimization of NTP-UFM (Eq. \eqref{eq:ufm}) with square loss.
    Assume infinitesimal step-size and spectral initialization $
\W(0)=e^{-\delta}\Ub\Rb^\top, \Hb(0)=e^{-\delta}\Rb\Vb^\top$
for partial orthogonal matrix $\Rb\in\R^{d\times r}$ ($\Rb^\top\Rb=\Id_r$) and initialization scale $e^{-\delta}$. Then, word and context embeddings at iteration $t$ evolve as
$
    \W(t) = \Ub\sqrt{\Sigmab}\sqrt{\Ab(t)}\Rb^\top $ and $\Hb(t) = \Rb\sqrt{\Sigmab}\sqrt{\Ab(t)}\Vb^\top
$
for $\Ab(t)=\diag{a_1(t),\ldots,a_r(t)}$ with
\begin{align}\label{eq:a_i(t) exact}
    a_i(t)=\frac{1}{ 1 + ( \sigma_i e^{2\delta} -1) e^{-2\sigma_i t}}
    , ~i\in[r].
\end{align}
Therefore, the $i$-th concept converges ($a_i(t)\rightarrow 1$) with  exponential rate of $2\sigma_i$: concepts corresponding to larger singular values are learned first.
\end{theorem}

{{Proof and justification of the spectral initialization are in App. \ref{app:saxe_connections}.} Our key insight is bridging the UFM with \citet{saxe2013exact}'s study of closed-form learning-dynamics of two-layer linear nets.

To empirically verify this in natural language, we track concept learning through confusion matrices at different training timestamps. Since  NTP performs soft-labeled classification where contexts can have multiple valid next-tokens, we define "effective classes" as the distinct sets of possible next-tokens (i.e., support sets) shared by at least one context in the dataset. For the dataset Fig.~\ref{fig:motivation}C, there are 9 such classes. Fig.~\ref{fig:concept_figures}C shows $9\times 9$ confusion matrices at four training timestamps from a 2-layer transformer; entry $(c,c')$ indicates how many contexts from class $c$ are predicted to belong to class $c'$. The diagonal entries show the number of correct classifications, with the total number of contexts per class shown after the slash.
We observe that certain effective classes naturally correspond to semantic categories, as contexts sharing semantic meaning (e.g., describing animal attributes) tend to share the same support set (e.g., animal names). We annotate these associations with labels ("Animals," "Plants") in brackets on the confusion matrix axes.

The visualization reveals that the model first learns classes with coarse semantic associations (e.g., "Animals" vs. "Plants" at step 20), then progressively distinguishes classes with finer-grained semantic associations (e.g., "Lives in water (fish)" by step 40). 
This progression reflects that coarse semantic structures, corresponding to concepts with larger singular values, are learned earlier in training than finer-grained ones that require combining more concept dimensions.

{To isolate the essential mechanism driving this progressive emergence, we identify the minimal data structure required: class imbalance alone suffices. Fig.~\ref{fig:confusion_steps} in Appendix demonstrates this on imbalanced MNIST, where the model learns distinctions in order of their corresponding singular values. Analyzer vectors with larger singular values encode majority-majority distinctions (learned first), the middle singular value encodes the majority-minority distinction (learned next), and smaller singular values encode minority-minority distinctions (learned last). This confirms that the progressive emergence of concept dimensions, ordered by singular value magnitude, stems from the fundamental optimization geometry determined by $\smatbar$ rather than being specific to linguistic structure. However, while even minimal imbalance produces meaningful categorical distinctions, richer sparsity patterns in $\smat$, as naturally arise in the soft-labeled language setting, enable progressively richer semantic concepts to emerge from its SVD, as we show in Sec.~\ref{sec:experiments}.}

\section{Experimental Validation}\label{sec:experiments} 

\noindent\textbf{Datasets.} We construct datasets by subsampling from two distinct corpora: TinyStories \citep{eldan2023tinystories} and WikiText-2 \citep{merity2016pointer}. For both, we fix vocabulary size $V=1$k and extract $m=10$k contexts by sampling frequent sequences of lengths 2--6 from the original corpora, creating \emph{Simplified TinyStories} and \emph{Simplified WikiText}. 

\noindent\textbf{Models.} To find semantics in models trained on controlled data, we train decoder-only transformers of varying sizes (6 or 12 layers, 1024 hidden dimensions) on Simplified datasets using AdamW until convergence {(see Appendix~\ref{sec:train_tf_detail})} for training details). We also analyze pretrained models starting with GPT-2-medium. To validate findings across architectures, scaling, and multi-language settings, we also analyze Qwen-7B (7B parameters with 4096-dimensional embeddings) \citep{bai2023qwen}. Finally, while we focus primarily on NTP loss, the underlying structure of context-to-next-token co-occurrence is also present in masked language modeling (MLM) loss, where the context consists of surrounding tokens and the model is trained to predict randomly masked tokens within the sequence. To verify our analysis applies to here as well, we experiment with BERT-base \citep{devlin2018bert}. }

\subsection{Methodology}\label{sec:methodology}
Our approach differs depending on whether we have access to the training corpus. \emph{Simplified datasets:} We explicitly construct  $\smatbar$ and compute word and context analyzer vectors for each concept via SVD (Sec.~\ref{sec:svd}). \emph{Pretrained models:} We do not have access to their pretraining corpora, making it infeasible to construct $\smatbar$ and compute its SVD. We thus resort to an approximation: we extract the decoder matrix $\W$, then use Arnoldi decomposition to compute their left singular vectors $\Ub$, which serve as proxies for word analyzer vectors. 

For both cases, once we have analyzer vectors, we apply orthant-based clustering (Sec.~\ref{sec:orthant}). We combine the top-$p$ concepts $\{1,\ldots,p\}$ in successive order, starting from the most dominant. We find that choosing $p$ as small as 5 to 7 and inspecting word or context membership across different sign configurations (i.e., orthants) typically suffices to reveal interpretable semantic patterns. 

To interpret semantics, we visualize top 40 words/contexts associated with each orthant. ``Association'' and ``top'' are evaluated per Defn.~\ref{defn:typicality}. We use word clouds with size proportional to typicality score. Symbol \# indicates token start.

\subsection{Results}
\label{sec:exp_findings}
\textbf{Individual concepts are not interpretable.}
We verify that not all individual concepts correspond to linguistically interpretable categories (Sec.~\ref{sec:svd}). Fig.~\ref{fig:concepts_not_interp} in App.  demonstrates this for Simplified TinyStories and  WikiText. This aligns with findings in the recent literature \citep{chersoni-etal-2021-decoding,piantadosi2024concepts,elhage2022toy}.


\begin{figure*}[t!]
\centering

\resizebox{0.82\textwidth}{!}{%
  \begin{minipage}{\textwidth}
    \noindent
    \begin{minipage}[t]{0.239\linewidth}\centering
      \textbf{\small Past-tense verb}\strut\\
      \includegraphics[width=\linewidth]{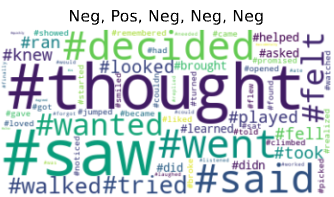}
    \end{minipage}%
    \hfill%
    \begin{minipage}[t]{0.239\linewidth}\centering
      \textbf{\small Present-tense verb}\strut\\
      \includegraphics[width=\linewidth]{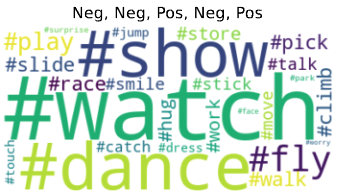}
    \end{minipage}%
    \hfill%
    \begin{minipage}[t]{0.239\linewidth}\centering
      \textbf{\small Preposition}\strut\\
      \includegraphics[width=\linewidth]{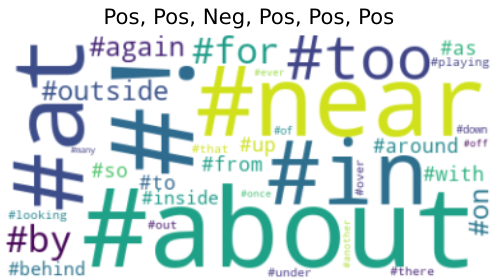}
    \end{minipage}%
    \hfill%
    \begin{minipage}[t]{0.239\linewidth}\centering
      \textbf{\small Proper names}\strut\\
      \includegraphics[width=\linewidth]{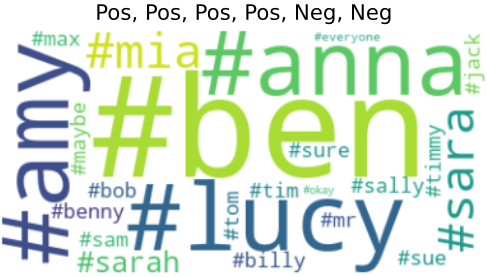}
    \end{minipage}%
    \vspace{0.2cm}

    \noindent
    \begin{minipage}[t]{0.239\linewidth}\centering
      \textbf{\small Month}\strut\\
      \includegraphics[width=\linewidth]{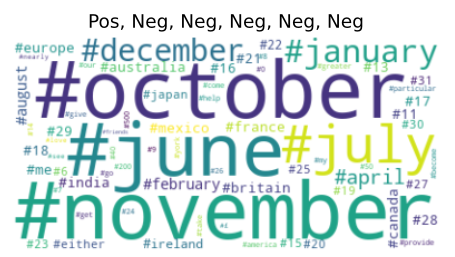}
    \end{minipage}%
    \hfill%
    \begin{minipage}[t]{0.239\linewidth}\centering
      \textbf{\small Number}\strut\\
      \includegraphics[width=\linewidth]{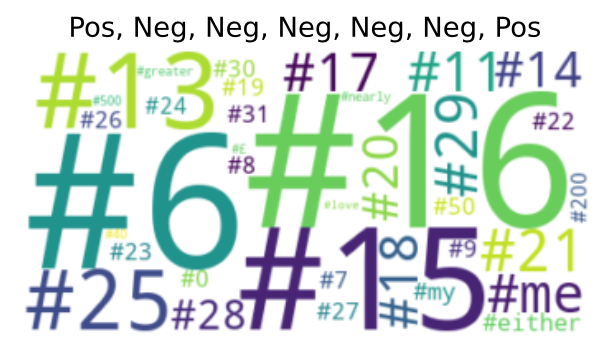}
    \end{minipage}%
    \hfill%
    \begin{minipage}[t]{0.239\linewidth}\centering
      \textbf{\small Unit}\strut\\
      \includegraphics[width=\linewidth]{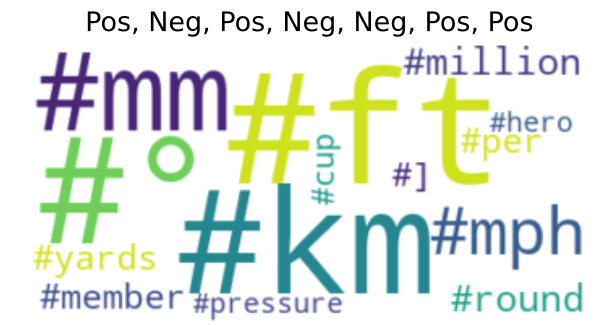}
    \end{minipage}%
    \hfill%
    \begin{minipage}[t]{0.239\linewidth}\centering
      \textbf{\small Preposition/Conjunction}\strut\\
      \includegraphics[width=\linewidth]{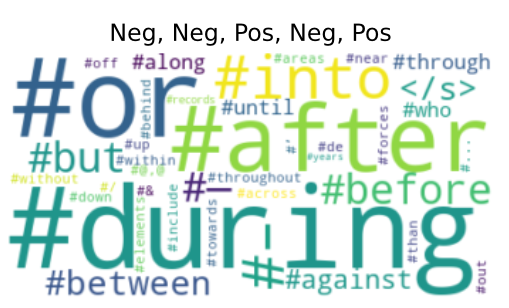}
    \end{minipage}%
  \end{minipage}%
}
\captionsetup{width=\textwidth}
\caption{Semantics from orthant-based clustering on $\smatbar$ in Simplified TinyStories \emph{(Top)} and Simplified Wikitext \emph{(Bottom)}. Semantic interpretations and combination configurations are shown in the title.
}
\label{fig:wordclouds_combined}
\end{figure*}

\begin{figure*}[!t]
\centering
\resizebox{0.82\textwidth}{!}{%
  \begin{minipage}{\textwidth}
    \begin{minipage}[t]{0.24\linewidth}
        \centering
        \textbf{\small Positive} \\
        \includegraphics[width=\linewidth]{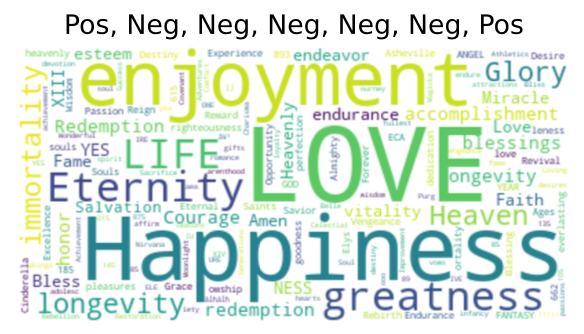}
    \end{minipage}
    \hfill
    \begin{minipage}[t]{0.24\linewidth}
        \centering
        \textbf{\small Negative} \\
        \includegraphics[width=\linewidth]{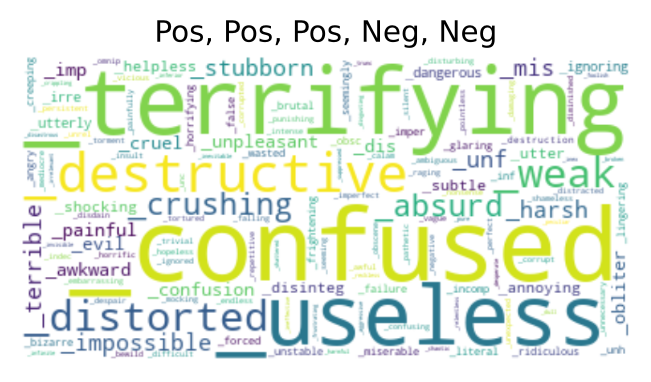}
    \end{minipage}
    \hfill
    \begin{minipage}[t]{0.24\linewidth}
        \centering
        \textbf{\small Chemical} \\
        \includegraphics[width=\linewidth]{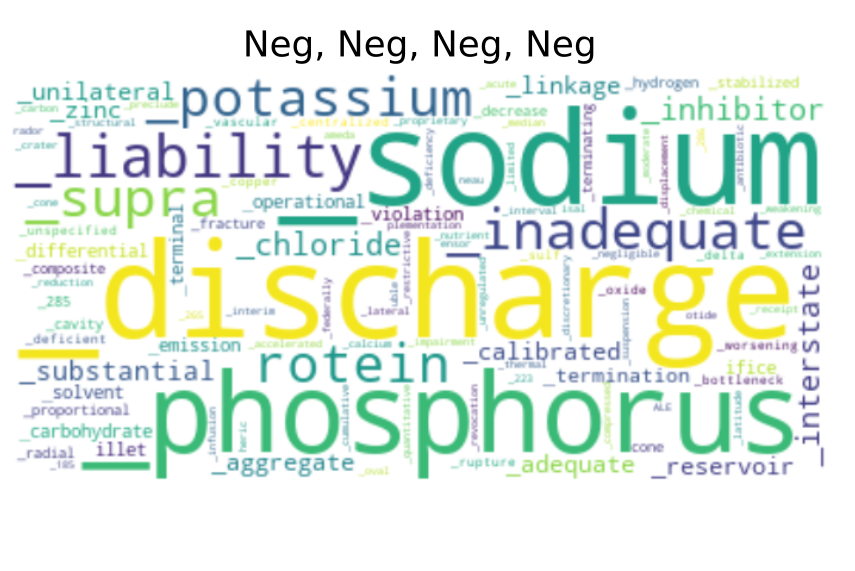}
    \end{minipage}
    \hfill
    \begin{minipage}[t]{0.24\linewidth}
        \centering
        \textbf{\small Medical} \\
        \includegraphics[width=\linewidth]{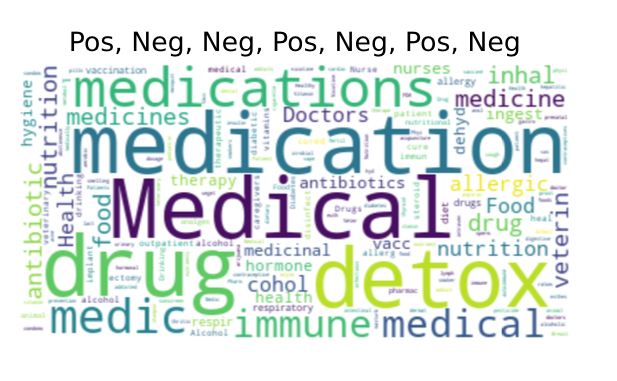}
    \end{minipage}

    \vspace{0.15cm}

    \begin{minipage}[t]{0.24\linewidth}
        \centering
        \textbf{\small Cooperation} \\
        \includegraphics[width=\linewidth]{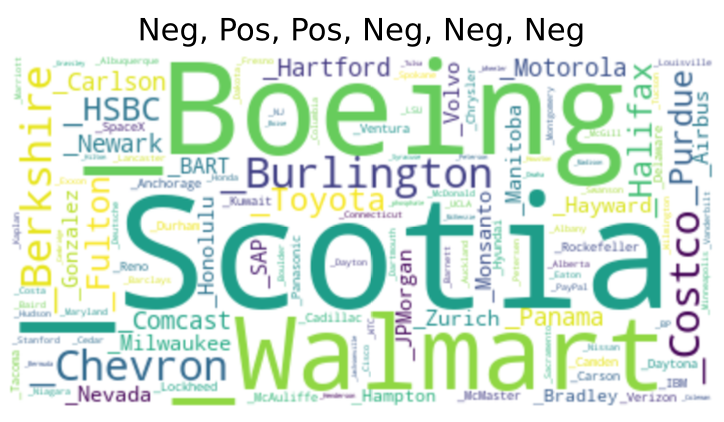}
    \end{minipage}
    \hfill
    \begin{minipage}[t]{0.24\linewidth}
        \centering
        \textbf{\small Digital entertainment} \\
        \includegraphics[width=\linewidth]{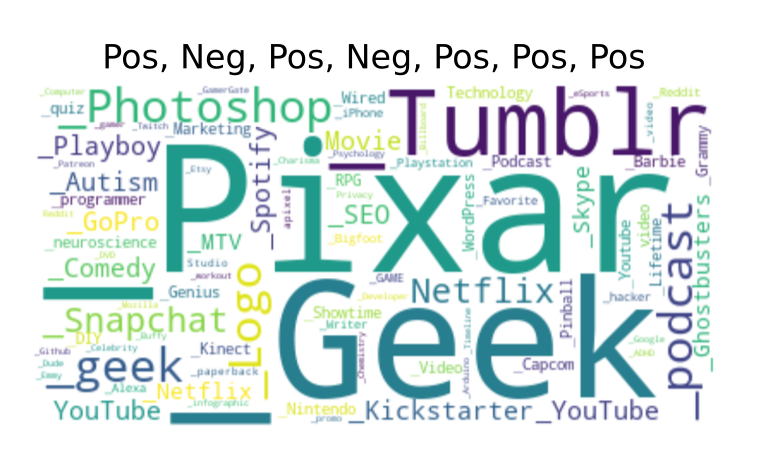}
    \end{minipage}
    \hfill
    \begin{minipage}[t]{0.24\linewidth}
        \centering
        \textbf{\small Cultural} \\
        \includegraphics[width=\linewidth]{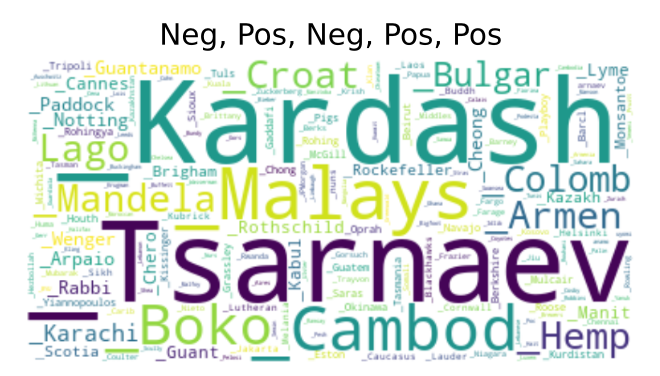}
    \end{minipage}
    \hfill
    \begin{minipage}[t]{0.24\linewidth}
        \centering
        \textbf{\small Societal} \\
        \includegraphics[width=\linewidth]{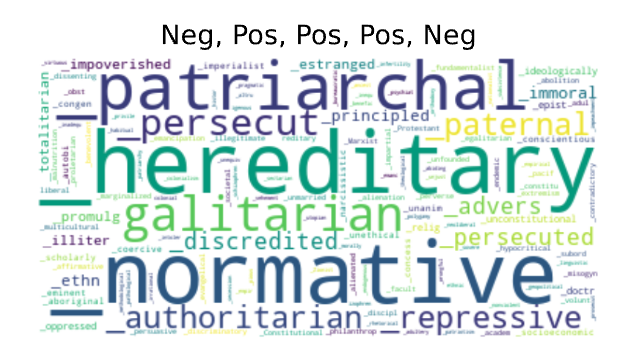}
    \end{minipage}
  \end{minipage}%
}
\captionsetup{width=\textwidth}
\caption{Semantics identified by orthant-based clustering on GPT-2’s word embeddings.}
\label{fig:pretrained_w}
\end{figure*}

\noindent\textbf{Combining concepts with sign configurations reveals semantics.} We verify that combining (non-interpretable) individual concepts reveals interpretable semantics via orthant-based clustering (Sec.~\ref{sec:orthant}). Fig.~\ref{fig:wordclouds_combined} shows human-interpretable semantic categories identified by orthant-based clustering in Simplified TinyStories and WikiText. In Fig.~\ref{fig:hierarchy} in Appendix, we further demonstrate the hierarchical semantic structure that emerges as we combine more concepts, illustrated through verb forms organized by grammatical properties. Starting with only the top concept ($k=1$), the categories remain broad and difficult to interpret. However, as we incorporate additional concepts, increasingly refined semantic categories emerge, ultimately distinguishing between different verb forms  (past tense, present tense, and present continuous).
\\
\indent Orthant-based clustering can be viewed as instance of spectral clustering \cite{spectral_clustering}. We thus also explore k-means spectral clustering over concept subspaces in App.~\ref{sec:app-kmeans}. While this also reveals interpretable categories, orthant-based clustering offers advantages: (1) It provides a direct link between optimization and categories by explicitly interpreting sign patterns of SVD dimensions as semantic indicators. (2) It has built-in selectivity: not all $2^p$ potential orthants yield semantically coherent groupings—only a meaningful subset does. (3) This control over $p$ enables semantic interpretation of membership and typicality, and allows exploration of different semantic granularities.

\begin{figure*}[t!]
    \centering
    \includegraphics[width=1\linewidth]{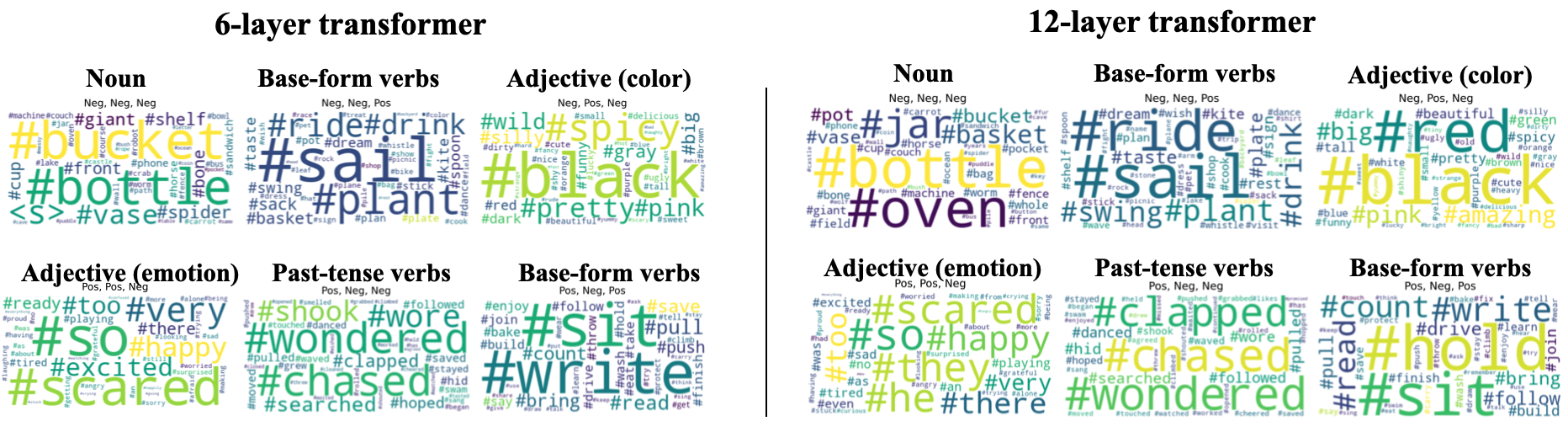}
    \captionsetup{width=\textwidth}
    \caption{
    Semantics from transformers of different depth, both trained on Simplified TinyStories. }
    \label{fig:tf_compare}
\end{figure*}

\begin{figure*}[t!]
\centering
\resizebox{0.82\textwidth}{!}{%
  \begin{minipage}{\textwidth}
    \noindent
    \begin{minipage}[t]{0.235\linewidth}\centering
      \includegraphics[width=\linewidth]{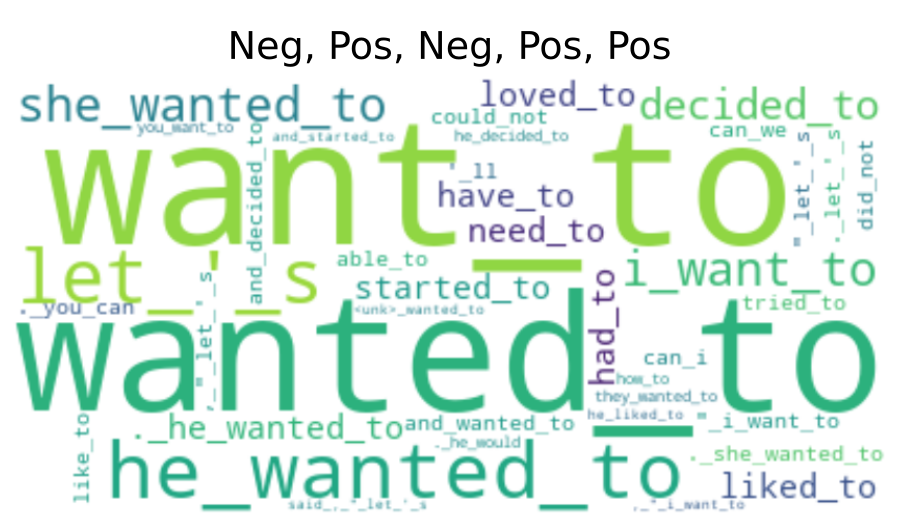}
    \end{minipage}%
    \hfill%
    \begin{minipage}[t]{0.235\linewidth}\centering
      \includegraphics[width=\linewidth]{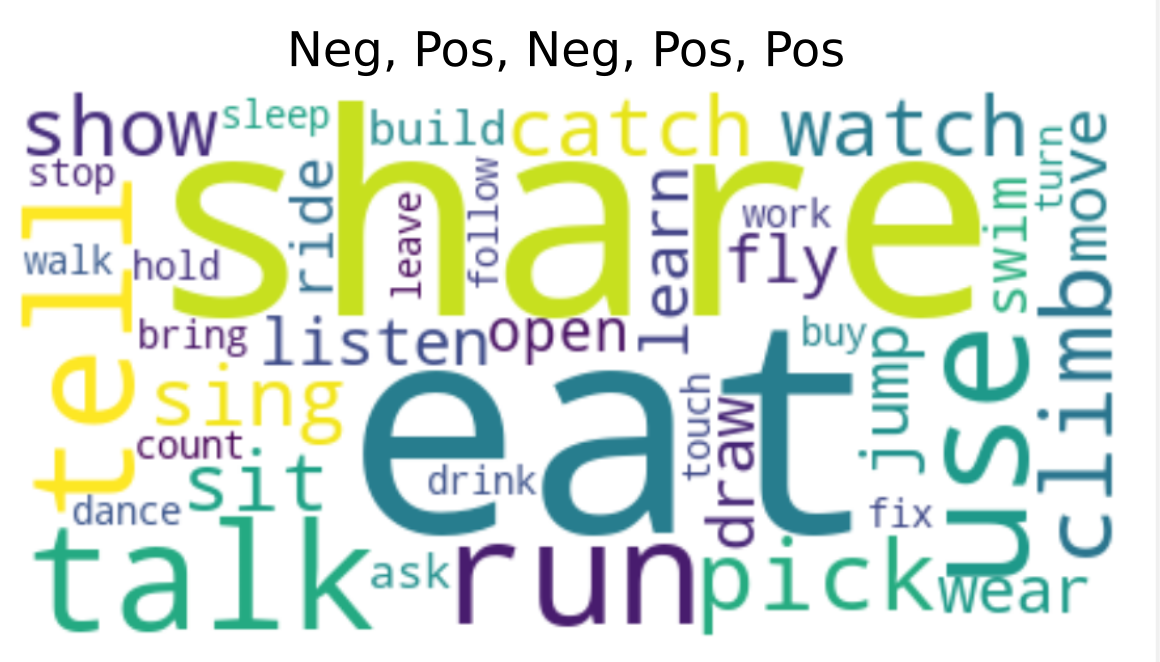}
    \end{minipage}%
    \hfill%
    \rule{0.5pt}{0.12\linewidth}
    \hfill%
    \begin{minipage}[t]{0.235\linewidth}\centering
      \includegraphics[width=\linewidth]{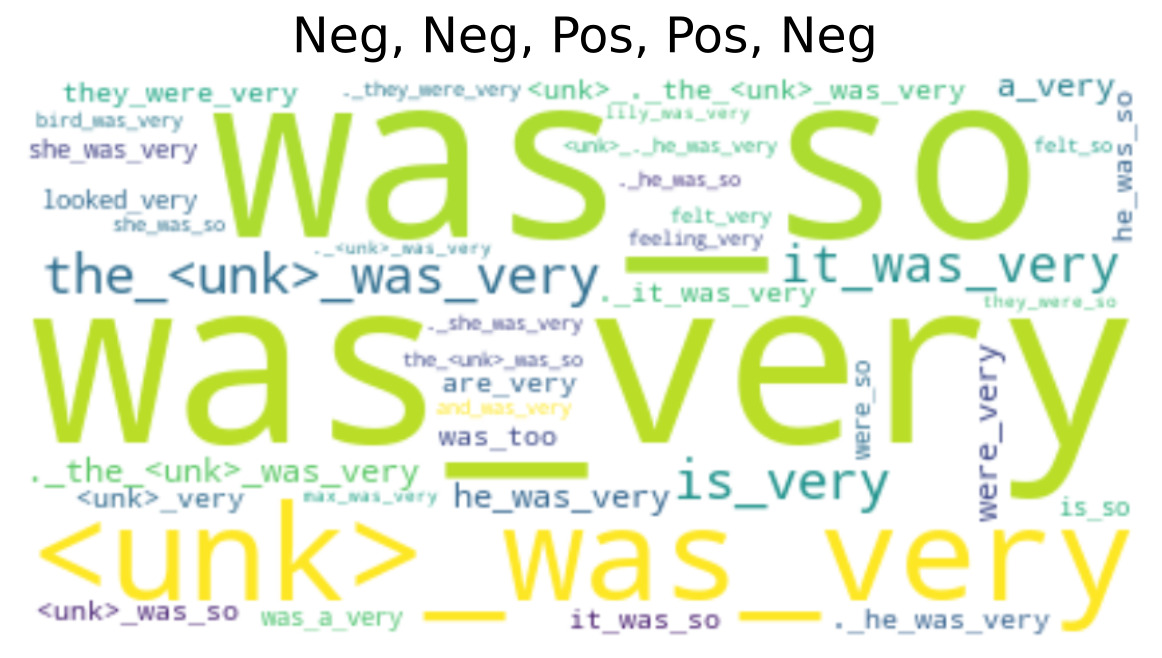}
    \end{minipage}%
    \hfill%
    \begin{minipage}[t]{0.235\linewidth}\centering
      \includegraphics[width=\linewidth]{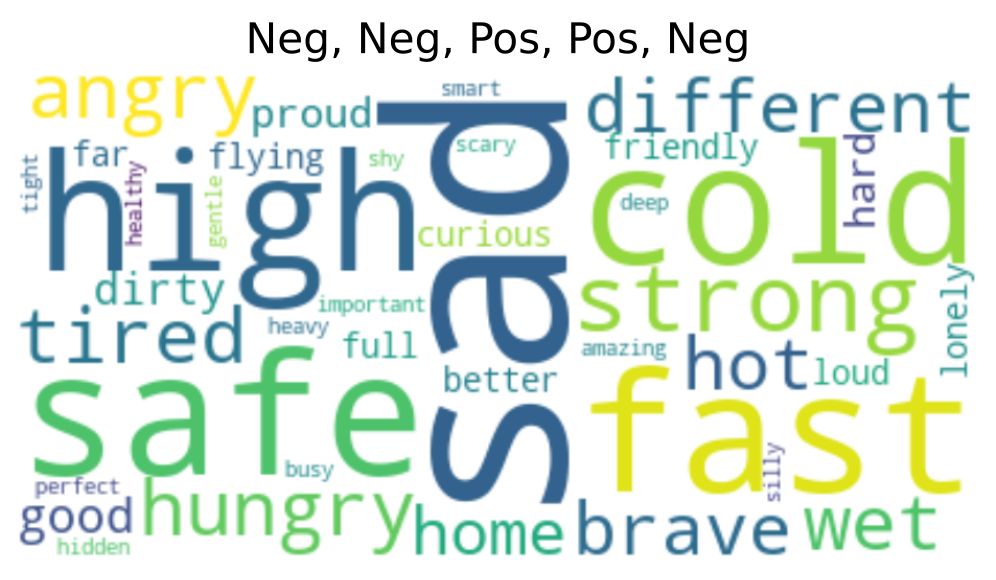}
    \end{minipage}%
  \end{minipage}%
}
\captionsetup{width=\textwidth}
\caption{Context/word semantics share identical configurations. Left and right pairs illustrate the ``verb+to infinitive'' and ``was so/very + adj'' structures, confirming semantic alignment between contexts and their in-support words.}
\label{fig:hwpair_ts}
\end{figure*}

\noindent\textbf{Relation of words and context semantics.} 
Recall from Secs.~\ref{sec:svd},\ref{sec:orthant} that semantic clustering of both word and context embeddings is guided by the SVD of the same data-sparsity matrix $\smatbar$. This coupling predicts that words and contexts sharing semantic properties should cluster together in embedding space. Fig.~\ref{fig:motivation}A illustrated this phenomenon in a simple setting, where plant-related words and contexts clustered separately from animal-related ones. We now demonstrate  this in more complex datasets: We apply orthant-based clustering to words and contexts using the same signature pattern across a subset of concept dimensions extracted from Simplified TinyStories. Fig.~\ref{fig:hwpair_ts} shows that contexts with modals like ``want/like/love/have to'' align with clusters of infinitive verbs; intensifiers like ``was so/very'' correspond to descriptive adjectives. \emph{This duality reflects how contexts encode meaning both via their surface words and via the semantics of likely next tokens}—a view naturally supported by the predictive nature of NTP. 

\noindent\textbf{Semantics in pretrained models.}
We find that simplified datasets primarily reveal grammatical/syntactic structure due to limited size and context diversity. To verify that richer sparsity induces richer semantics, we here analyze pretrained models. 
In {GPT-2} (Fig.~\ref{fig:pretrained_w}), orthant-based clustering recovers categories such as medical terms, emotional tone, entertainment, and political figures.  In {BERT} (Fig.~\ref{fig:bert_orthant} in App.), we find clusters reflecting numerical patterns (e.g., 3-digit numbers), action verbs, administrative phrases, and morphological forms. Applying the same method to {Qwen-7B} yields UI phrases, programming keywords, numeric tokens, functional Chinese expressions, and so on. Qwen requires combining more concepts (7–11 vs. 5–7 for GPT-2) to isolate coherent groups (Table~\ref{tab:qwen} in App.), suggesting more distributed semantic encoding, {potentially reflecting its multilingual training.}  

\noindent\textbf{Consistent structure across models.} Our analysis predicts that, as long as an architecture is sufficiently expressive for the UFM assumption to hold, the geometry of learned semantics should depend on data patterns rather than specific architectural details. We verify this by training two transformers with different depths (6 vs. 12 layers) on the same Simplified TinyStories dataset. After training, we extract the decoder matrix $\W$ from each model and compute its top three left singular vectors. Applying orthant-based clustering, both models yield nearly identical semantic groupings (Fig.~\ref{fig:tf_compare}).

\section{Related Work}
\label{sec:related_work_main}
We bring together, in the context of NTP-trained language models, two classical ideas: (1) semantics emerge from linear relationships between vector representations of textual inputs, and (2) learned-representations geometries are tied to language statistics via count matrices. App. \ref{sec:related_work_full} for details. 

Extracting semantics from distributed representations was popularized through word embedding research \cite{mikolov2013efficient}, with approaches including semantic axes \cite{An2018SemAxis, Fast2016Empath} and word analogies \cite{mikolov2013distributed, levy_semantics}. In LLMs,  work has focused on dictionary learning \cite{turner2023activation, li2024inference}, geometric clustering \cite{coenen2019visualizing, hewitt2019designing}, and linear probing \cite{hewitt2019structural, marks2023geometry}. Unlike  post-hoc analyses, \emph{we analyze, from an optimization perspective, how semantics emerge as a consequence of training.}
While \citet{park2023linear, park2024geometry} have explored similar questions, here  we directly analyze NTP optimization, rather than relying on a generative model for concept generation and its link to words/contexts.

Classical work has shown that applying factorization methods to co-occurrence matrices conveys semantic information \cite{landauer1997latent, levy_semantics}, with various hand-engineered choices for both the matrix representation (e.g., word-document, word-word, PMI, PPMI) and factorization technique (SVD, NNMF \cite{lee1999learning, ding2006nonnegative}, sparse dictionary learning \cite{murphy2012learning, faruqui2015sparse}). 
We demonstrate, for the first-time, that gradient-based optimization itself determines both the matrix representation ($\smatbar$) and factorization method (SVD) used to encode semantics. 

Our work connects NTP-UFM to the closed-form framework of \citet{saxe2013exact, saxe2019mathematical} for linear neural networks. While their work focuses on general cognitive development with orthogonal inputs, we provide a new practical instantiation for language modeling optimization and validate  theory findings on real data and architectures.

\section{Conclusion}
\label{sec:conclusion}
As LLM capabilities continue to advance, understanding their inner workings is crucial. We study how language-modeling optimization guides the geometry of word and context representations to organize around latent semantic information. Without explicitly computing it, the model encodes such information in the SVD factors of an implicit data-sparsity matrix. As an early contribution in this direction, our analysis motivates several future directions, some of which we outline in Sec. \ref{sec:limitations}.

\newpage
\section{Limitations}\label{sec:limitations}
\textbf{(1)} While the assumption $d\geq V$ allows already for rich geometric structures, most practical models use $d<V$. {We hypothesize that during NTP training, models learn to represent the $d$ most significant concepts, corresponding to the largest singular values of $\smatbar$. While NTP-UFM offers preliminary support of this hypothesis, a rigorous theoretical and experimental analysis requires separate future investigation. Encouragingly, since semantic categories arise from concept combinations (Sec.~\ref{sec:orthant}), even a reduced set of core concepts could enable rich semantic representations.}
\textbf{(2)} Our analysis assumes sufficient expressivity and training time, which may become increasingly relevant with improved compute or in data-limited regimes, but relaxing them 
or experimentally validating their applicability in the spirit of our GPT-2/Qwen experiments
is important.
\textbf{(3)} 
 Existing theoretical work on transformer optimization dynamics is limited to shallow models (e.g. one-layer)  and overly strict data-generation assumptions that limit capturing meaningful semantic information. On the other hand, the abstraction of unconstrained features offers mathematical tractability, already useful insights, and has the attribute of leading to predictions insensitive to the specifics of the architecture (Sec. \ref{sec:exp_findings}). Future work can explore merging the two paradigms of analysis.
\textbf{(4)} While our approach differs from concurrent efforts to explain semantic emergence in LLMs via probing or sparse-dictionary learning, future work could explore connections and potential applications to model improvement techniques.

\section{Acknowledgments}
This work was funded by the NSERC Discovery Grant No. 2021-03677, the Alliance Grant ALLRP 581098-22, and an Alliance Mission Grant. The authors also acknowledge use of the Sockeye cluster by UBC Advanced Research Computing.

\bibliography{refs,transformers,refs_NC,bib_extra,compbib} 


\newpage
\onecolumn
\appendix
\onecolumn

\appendix

\begin{figure*}
\vspace{0.2in}
    \centering
    \includegraphics[width=0.8\linewidth]{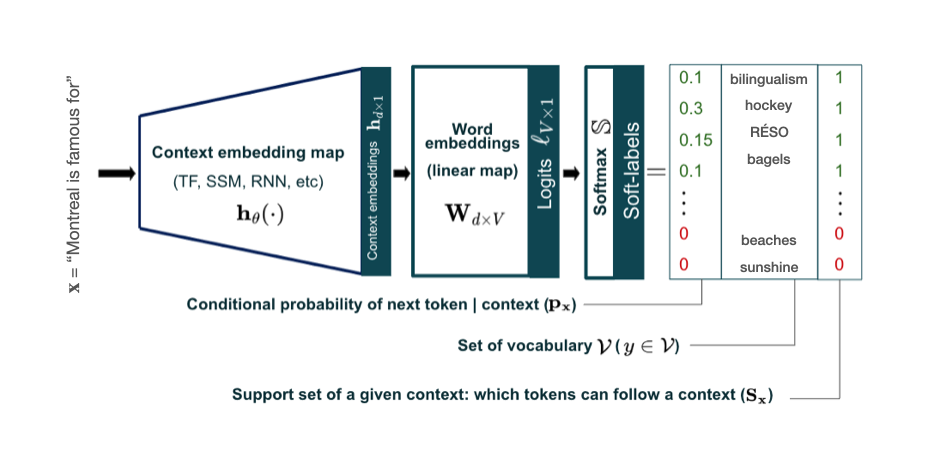}
    
    \captionsetup{width=\textwidth}
    \caption{Illustration of setup and terminology.}
    \label{fig:model}
\end{figure*}

\section{Paper Organization}\label{sec:related_work_full}

The main paper is organized as follows: Sec.~\ref{sec:background_main} introduces background on our analytical model and the geometry of embeddings. Our main analytical contributions and minimal synthetic experiments are presented in Sec.~\ref{sec:concept geometry}, with subsections~\ref{sec:svd}--\ref{sec:rate} each detailing one of the three contributions (C1-C3). Sec.~\ref{sec:experiments} provides qualitative studies on real-world data and pretrained models that validate our findings. Sec.~\ref{sec:related_work_main} contextualizes our work within the broader literature, with comparisons to closely related work integrated throughout the paper. We conclude in Sec.~\ref{sec:conclusion} and discuss limitations and future directions in Sec.~\ref{sec:limitations}.

The appendix is organized as follows: Sec.~\ref{sec:related_work_full} provides an extended discussion of related work (a condensed version appears in the main paper). Sec.~\ref{sec:proofs} includes details on the formalization of concept representations discussed in Sec.~\ref{sec:svd} and the proof of Theorem~\ref{thm:main} from Sec.~\ref{sec:rate}. Additional experimental results deferred due to space constraints appear in Sec.~\ref{app:additional_exp}.

\section{Extended Discussion on Related Work}\label{sec:related_work_full}

This section provides an extended discussion of the related work summarized in Sec.~\ref{sec:related_work_main}.

    \noindent\textbf{Semantics from Distributed Representations.}
The idea of extracting semantics from distributed representations, while rooted in earlier works \cite{bengio2000neural}, was popularized through word embedding research \cite{mikolov2013efficient}. Two classically popular approaches include: (i) constructing semantic axes from centroids of pole words \cite{An2018SemAxis, Fast2016Empath}, evolving from lexicon-based sentiment analysis \cite{Taboada2011, Hu2004}, and (ii) investigating semantic patterns through ``word analogies,'' which reveal that semantic relationships manifest as linear directions in embedding space \cite{mikolov2013distributed, levy_semantics, drozd-etal-2016-word, ethayarajh-etal-2019-towards, allen2019analogies}.
In the context of modern LLMs, word analogies have been empirically observed \cite{rezaee-camacho-collados-2022-probing, wijesiriwardene-etal-2024-relationship}. However, recent focus has shifted to dictionary learning techniques for distilling interpretable semantics, e.g., \cite{turner2023activation, li2024inference}. 
Alternative approaches include geometric studies showing how linguistic features cluster in embedding spaces \cite{coenen2019visualizing, hewitt2019designing}. Simultaneously, supervised methods such as linear probing have been employed to extract semantic and syntactic information from modern LLMs \cite{alain2017understanding, hewitt2019structural,marks2023geometry}, relying on the same principle of linear representation found in word analogies.
Unlike these works that empirically analyze trained models post-hoc, \emph{our goal is not to introduce a state-of-the-art method for extracting semantics, but rather to understand, from an optimization viewpoint, how semantics emerge as a consequence of next-token prediction.}  While \citet{park2023linear, park2024geometry} have explored similar questions, here  we directly analyze NTP optimization, rather than relying on a generative model for concept generation and its link to words/contexts.

\noindent\textbf{Geometry from Co-occurrence Statistics and Matrix Factorization.}
The idea that factorizations of co-occurrence matrices convey latent semantic information dates back to latent semantic analysis of word-document co-occurrence matrices \cite{landauer1997latent, deerwester1990indexing} and later to pointwise mutual information matrices of word-word co-occurrences \cite{levy, levy_semantics}. Beyond SVD, alternative factorization techniques such as non-negative matrix factorization \cite{lee1999learning, ding2006nonnegative} and sparse dictionary learning \cite{murphy2012learning, faruqui2015sparse} have been applied to various transformations of co-occurrence data, including those addressing PMI sparsity (e.g., PPMI) or using probability smoothing \cite{levy_semantics}.

More broadly, vector space representations emerge from two approaches: (i) traditional count-based models using co-occurrence matrices, and (ii) prediction-based models that optimize token prediction objectives like Skip-gram or NTP. \citet{levy,glove} demonstrate that these approaches are fundamentally similar, as both probe underlying corpus co-occurrence statistics. \citet{zhao2024implicit} have extended this insight to causal NTP models with a key distinction: the co-occurrence matrix between contexts and words is inherently sparse, unlike the dense word-word co-occurrences in prior work. This distinction makes relevant recent advances in implicit regularization \cite{soudry2018implicit} and neural collapse geometries \cite{papyan2020prevalence,mixon2020neural}. 

\noindent\textbf{Relation to \citep{zhao2024implicit}.} Our work extends \citet{zhao2024implicit}  by addressing the three open questions Q1-Q3 discussed in our Introduction. {In doing so, we provide crucial semantic interpretation to prior NC-based representation analyses (including those for imbalanced one-hot settings \cite{fang2021exploring,seli}). Furthermore, by connecting our model to the closed-form analysis of linear neural network learning dynamics \cite{saxe2013exact}, we rigorously demonstrate how singular values govern the order of concept acquisition during training.mine the learning order during training. 

\noindent\textbf{Relation to \citet{saxe2019mathematical}.} Finally, while drawing conceptual similarities to \citep{saxe2019mathematical}, our work differs in its motivation, interpretation, and scope of results. Their focus is on general cognitive development using an abstract two-layer linear neural network with orthogonal inputs. In contrast, we specifically analyze how semantic and grammatical concepts emerge from natural language data through NTP optimization. Methodologically, we realize a connection and apply their framework to the NTP-UFM (which we interpret as a linear two-layer network where the input dimension equals the number of input examples), a tractable model for expressive large architectures. This provides a practical instantiation for their otherwise restrictive orthogonal-input assumptions.  We also corroborate our findings with experiments on real datasets and architectures, a crucial distinction from their purely theoretical work.

\onecolumn
\section{Derivations and Proofs}\label{sec:proofs}

\subsection{Concepts in Embedding Space}\label{app:concept geometry app}
This section includes derivations of the formulas of \wco and \cco representations of Sec. \ref{sec:svd}. 
Recall the centered data-sparsity matrix $\Sbar$ and  its SVD 
\[
\Sbar=\Ub\Sigmab\Vb^\top,\quad \text{where}\quad \Ub\in\R^{V\times r}, \Sigmab\in\R^{r\times r}, \Vb\in\R^{m\times r}\quad\text{and}\quad \Ub^\top\Ub=\Vb^\top\Vb=\Id_r\,,
\]
with singular values $\Sigmab=\diag{\sigma_1,\ldots,\sigma_r}$ that are ordered:
\[
\sigma_1\geq \sigma_2 \geq \ldots \geq \sigma_r >0\,.
\]
 We interpret the columns $\ub_k\in\R^V, \vb_k\in\R^m, k\in[r]$  of $\Ub, \Vb$  as {word} and {context} {analyzer vectors} for {concept} $k$.
For each word $z\in\Vc$ and each \wco $k\in[r]$, the component $\ub_k[z]$ represents how \emph{present} or \emph{absent} is a word $z$ in context $k$. Respectively for contexts. Thus, we think of column dimensions of $\Ub$ as concepts that capture semantic categories. For completeness, we note that each word-analyzer vector $\ub_k$ has both positive and negative entries, as it is orthogonal to $\ones_V$ (which lies in the nullspace of $\smatbar$).
%

\textbf{Q:} How do we define {\wco} and {\cco} representations, i.e. $d$-dimensional representations of word and context analyzer vectors for various concepts?

Let $\W\in\R^{V\times d}$ and $\Hb\in\R^{d\times m}$ be the representations of words and contexts. We then define \textbf{\wco representations} $\ubd_k$ and  \textbf{\cco representations}  $\vbd_k$ for $k\in[r]$ as projections onto the spaces of word and context representations, respectively. Specifically, for projection matrices
\[
\Pb_\W=\W^\top(\W\W^\top)^{-\dagger}\W\qquad\text{and}\qquad
\Pb_\Hb=\Hb(\Hb^\top\Hb)^{-\dagger}\Hb^\top\,,
\]
let
\begin{align*}
    \ubd_k &= \Pb_\W\W^\top\ub_k\,
\qquad\text{and}\qquad
    \vbd_k &= \Pb_\Hb\Hb\vb_k\,.
\end{align*}
We can further simplify these by using the known SVD representation of $\W$ and $\Hb$ from Sec. \ref{sec:background_main}.
Using this representation (i.e. use $\W\leftarrow\Wmm$, $\Hb\leftarrow\Hmm$) we compute
$
\Pb_\W=\Rb\Rb^\top=
\Pb_\Hb,
$
where recall that $\Rb$ is a partial orthogonal matrix.
Thus,
\begin{subequations}\label{eq:word context concepts same}
\begin{align}
    \ubd_k &= \Rb\Rb^\top\Rb\sqrt{\Sigmab}\Ub^\top\ub_k=\sigma_k \Rb\eb_k=\W^\top\ub_k\,\\
    \vbd_k &= \sigma_k \Rb\eb_k = \Hb\vb_k = \sum_{j\in[m]}\vb_k[j]\cdot\hb_j\,.
\end{align}
\end{subequations}
We conclude that the $d$-dimensional representations of word and context analyzer vectors are the same. {This is intuitive since concepts are categories in a certain sense broader than words/contexts, where the latter can be thought of as realizations of the concept in the form of explicit constituents of natural language.
} We thus refer to $\ubd_k=\vbd_k=\Rb\eb_k$ as the {representation} of concept $k$. See Fig. \ref{fig:d_dim_vis} for a visualization.

Finally, observe from Eq. \eqref{eq:word context concepts same} that {the $k$-th concept representation is given by a weighted average of word or context embeddings with weights taken by the resppective context analyzer vectors.}

\begin{figure}[H]
    \centering
    \begin{minipage}{0.45\textwidth}
        \centering
        \begin{tikzpicture}
            \node[inner sep=0pt] (image1) at (0,0) {\includegraphics[width=0.7\textwidth]{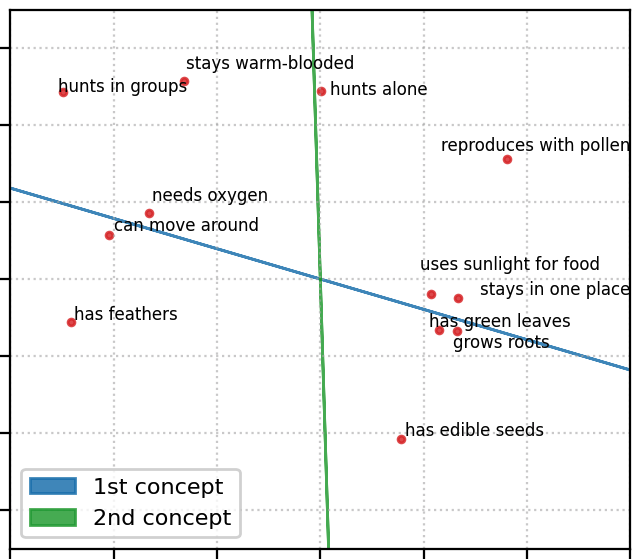}};
        \end{tikzpicture}
    \end{minipage}
    \begin{minipage}{0.45\textwidth}
        \centering
        \begin{tikzpicture}
            \node[inner sep=0pt] (image2) at (0,0) {\includegraphics[width=0.7\textwidth]{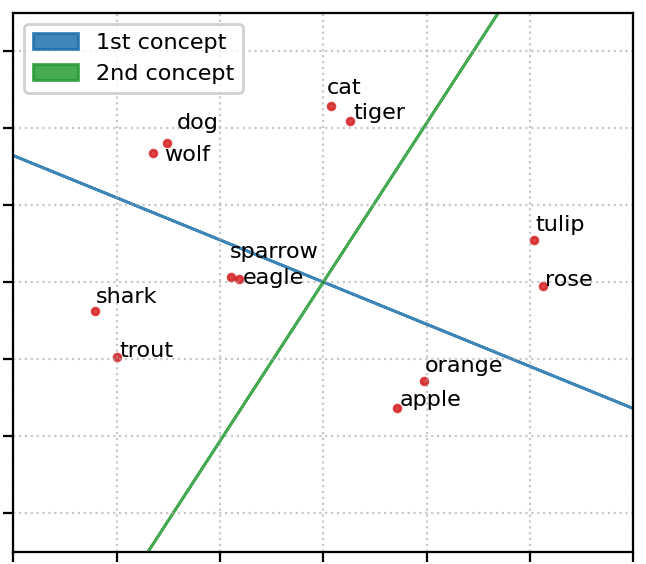}};
        \end{tikzpicture}
    \end{minipage}
\captionsetup{width=\textwidth}
\caption{Left: Context embeddings; Right: Word embeddings. In both plots, concept representations (blue, green) are computed using Eq. \eqref{eq:word context concepts same}. Projections onto these concept axes ($\vbd_1$, $\vbd_2$) quantify semantic relevance.  The first concept ($\vbd_1$, blue line) defines a spectrum from plant-related to animal-related elements, with animals (``hunts in groups'', ``dog'') projecting positively and plants (``uses sunlight for food'', ``apple'') projecting negatively along the same axis. The second concept ($\vbd_2$, green line) represents a continuum from non-mammalian to mammalian traits, with mammalian words and features (``cat'', ``stays warm-blooded'') having positive projections and non-mammalian ones (``trout'', ``has feathers'') having negative projections. (The concepts are not orthogonal in the plots because it is PCA projection to 2D space.) }
    \label{fig:d_dim_vis}
\end{figure}

\subsection{Proof of Theorem \ref{thm:main}}
\label{app:saxe_connections}

Consider the following square-loss NTP-UFM proxy:
\begin{align}\label{eq:UFM_L2}
    \min_{\W,\Hb}~~\left\{\sum_{j=1}^m\|\sbar_j-\W\hb_j\|^2=\|\Sbar-\W\Hb\|^2\right\}\,.
\end{align}
which fits logits $\W\Hb$ to the centered sparsity matrix $\Sbar$. For one-hot labels, this reduces to standard square-loss classification, which has shown competitive performance to CE minimization in various settings (\cite{hui2020evaluation,demirkaya2020exploring}). However, in our soft-label setting, the choice of loss function requires more careful consideration. While one could follow Glove's approach \cite{glove} by using $\log(\Pbb)$ instead of $\smatbar$, this creates issues with $\Pbb$'s zero entries. Since CE loss minimization leads $\W$ and $\Hb$ to factor $\smatbar$, we thus maintain $\smatbar$ as the target in \eqref{eq:UFM_L2}.

The neural-collapse literature has extensively studied Eq. \eqref{eq:UFM_L2} for one-hot $\sbar$, primarily in the balanced case (e.g., \cite{mixon2020neural,han2021neural,sukenik2023deep,tirer2022extended}) but recently also for imbalanced data (e.g., \cite{liu2024exploration,hong}). Most works focus on global minima of regularized UFM, with less attention to unregularized cases or training dynamics. While some landscape analyses provide partial answers about global convergence, they are limited to regularized cases and do not characterize dynamics. Even for square loss, where \citet{mixon2022neural,han2021neural} study training dynamics—notably \cite{han2021neural}'s analysis of the 'central path' in balanced one-hot cases—these results rely on approximations. Thus, a significant gap remains in understanding UFM training dynamics, even for simple balanced one-hot data with square loss.

By interpreting the UFM with square loss in Eq. \eqref{eq:UFM_L2} as a two-layer linear network with orthogonal inputs, we identify an unexplored connection to \citet{saxe2013exact,bach_saxe}'s analysis. They provide explicit characterization of gradient descent dynamics (with small initialization) for square-loss UFM. The key insight is rewriting \eqref{eq:UFM_L2} as $\sum_{j\in[m]}|\sbar_j-\W\Hb\eb_j|^2$ with orthogonal inputs $\eb_j\in\R^m$. This enables  application of their result, originally stated in \cite{saxe2013exact} and formalized in \cite{bach_saxe}. The following is a detailed (re)statement of Theorem \ref{thm:main}.

\begin{theorem}
    \label{thm:saxe} Consider gradient flow (GF) dynamics for minimizing the square-loss NTP-UFM \eqref{eq:UFM_L2}. Recall the SVD $\Sbar=\Ub\Sigmab\Vb^\top$. Assume weight initialization (see Fig. \ref{fig:convergence} for empirical justification of this form of spectral initialization)
\[
\W(0)=e^{-\delta}\Ub\Rb^\top \quad\text{and}\quad \Hb(0)=e^{-\delta}\Rb\Vb^\top
\]
for some partial orthogonal matrix $\Rb\in\R^{d\times r}$ ($\Rb^\top\Rb=\Id_r$) and initialization scale $e^{-\delta}$. Then the iterates   $\W(t),\Hb(t)$ of GF are as follows:
\begin{align}
    \W(t) = \Ub\sqrt{\Sigmab}\sqrt{\Ab(t)}\Rb^\top \qquad\text{and} \qquad \Hb(t) = \Rb\sqrt{\Sigmab}\sqrt{\Ab(t)}\Vb^\top
\end{align}
for $\Ab(t)=\diag{a_1(t),\ldots,a_r(t)}$ with
\begin{align}\label{eq:a_i(t) exact}
    a_i(t)=\frac{1}{ 1 + ( \sigma_i e^{2\delta} -1) e^{-2\sigma_i t}}
    , ~i\in[r].
\end{align}
Moreover, the time-rescaled factors $a_i(\delta t)$ converge to a step function as $\delta\rightarrow\infty$ (limit of vanishing initialization):
\begin{align}
    a_i(\delta t)\rightarrow \frac{1}{1+\sigma_i}\ones[{t=T_i}]+ \ones[{t>T_i}],
\end{align}
where $T_i=1/\sigma_i$ and $\ones[{A}]$ is the indicator function for event $A$. Thus, the $i$-th component is learned at time $T_i$ inversely proportional to $\sigma_i$.
\end{theorem}
\begin{proof}
    After having set up the analogy of our setting to that of \cite{saxe2013exact,bach_saxe}, this is a direct application of \citep[Thm.~1]{bach_saxe}. Specifically, this is made possible in our setting by: (i) interpreting the UFM with square-loss in \eqref{eq:UFM_L2} as a two-layer linear network (ii) recognizing that the covariance of the inputs (which here are standard basis vectors $\eb_j, j\in\R^m$) is the identity matrix, hence the (almost) orthogonality assumption (see \citep{saxe2013exact} and \citep[Sec.~4.1]{bach_saxe}) holds.
\end{proof}

The result requires initializing word/context embeddings aligned with the SVD factors of the data-sparsity matrix. While this might appear as a strong assumption, \cite{saxe2013exact,saxe2019mathematical} conjectured and verified experimentally that the characterization remains qualitatively accurate under small random initialization. Our experiments with the data-sparsity matrix confirm this - Fig. \ref{fig:rate} shows the singular values of the logit matrix during training closely follow the predicted exponential trend in Eq. \eqref{eq:a_i(t) exact}. This reveals that dominant singular factors, corresponding to primary semantic concepts, are learned first. In the limit $t\rightarrow\infty$, the theorem shows convergence to:
\begin{align}\label{eq:Winf}
\W(t) \rightarrow \Winf:=\Ub\sqrt{\Sigmab}\Rb^\top \qquad\text{and} \qquad \Hb(t) \rightarrow \Hbinf:=\Rb\sqrt{\Sigmab}\Vb^\top,.
\end{align}
This aligns with the regularization-path analysis of NTP-UFM with CE loss, discussed in Sec. \ref{sec:background_main}, where normalized quantities converge as $\la\rightarrow0$:
\[
\Wbar(\la) \rightarrow \Ub\sqrt{\Sigmabbar}\Rb^\top \qquad\text{and} \qquad \Hbbar(\la) \rightarrow \Rb\sqrt{\Sigmabbar}\Vb^\top.
\]
\begin{figure}[t]
   \centering
   \includegraphics[width=\textwidth]{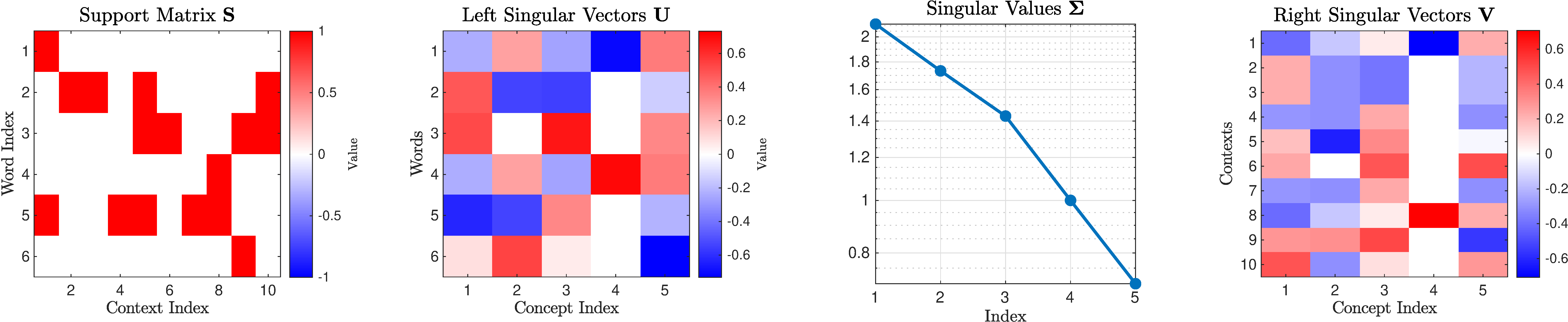}
   \vspace{1em}
   \hspace{-0.38in}\includegraphics[width=1.1\textwidth]{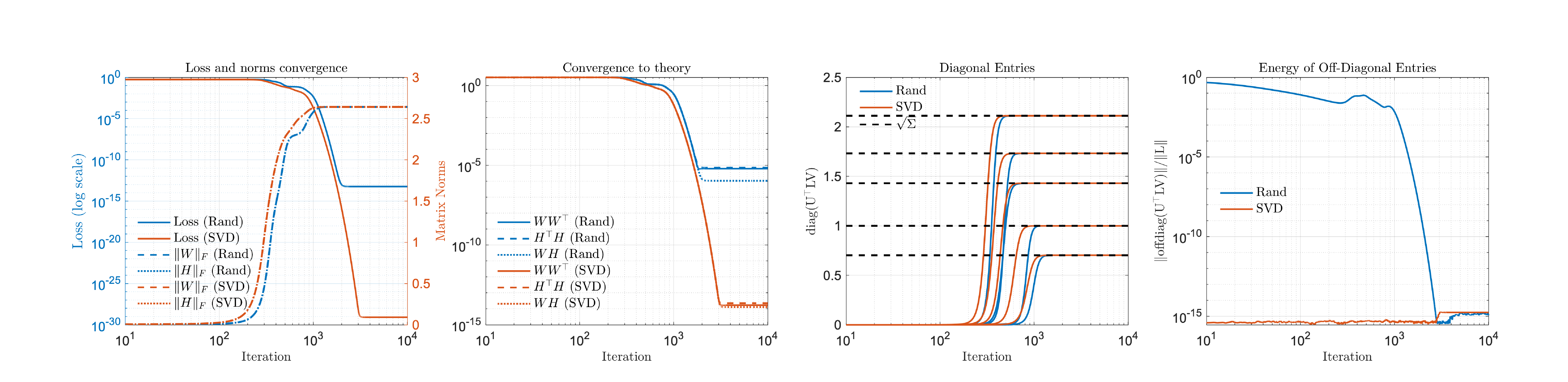}
\captionsetup{width=\textwidth}
   \caption{  \textbf{(Top)} Support matrix and SVD factors of centered support matrix for a synthetic example. \textbf{(Bottom)} Training dynamics of GD minimization of NTP-UFM with square loss (Eq. \eqref{eq:UFM_L2} for two initializations: (i) SVD: initialize $\W$ and $\Hb$ as per Thm. \ref{thm:saxe} for $\delta=8$. (ii) Rand: intialize $\W$ and $\Hb$ random Gaussian scaled to match the norm of SVD initialization. Dynamics with the two initialization are shown in red (SVD) and blue (Rand), respectively. Qualitatively the behavior is similar. \emph{Left:} Training loss and norms of paramteres. \emph{Middle-Left:} Convergence of word and context gram-matrices and of logits to the theory predicted by Thm. \ref{thm:saxe}. \emph{Middle-Right:} Convergence of singular values of logit matrix to those of $\Sigmab$ (see Thm. \ref{thm:saxe}). \emph{Right:} Projection of logits to subspace orthogonal to $\Ub$ and $\Vb$; Logits with Rand initialization initially have non-zero projection but it becomes zero as training progresses.}
   \label{fig:convergence}
\end{figure}

\clearpage

\section{Additional Experimental Results} \label{app:additional_exp}

\subsection{Minimal One-Hot Imbalanced Setting}\label{sec:NC}

In Sec.~\ref{sec:rate}, we identified class imbalance as the minimal requirement for progressive semantic emergence. Here, we provide detailed analysis of this minimal setting.

The emergence of semantically meaningful concepts stems from the structure of labels encoded in $\smat$ (or its centered version $\smatbar$) across different contexts. Consider the minimal case: each context has exactly one next-token, with contexts distributed equally across the vocabulary. Here, $\smat$ can be rearranged as $\Id_V\otimes\ones_{m/V}^\top$, yielding trivial concepts where all singular directions contribute equally. This setting corresponds to balanced one-hot classification studied in the neural collapse literature, where last-layer embeddings and weights form highly symmetric aligned structures \cite{NC}, reflecting the symmetric nature of $\smat$ with no interesting conceptual structure.

However, such balanced settings are unrealistic for language modeling. As seen in the controlled and real experiments in Secs.~\ref{sec:concept geometry} and~\ref{sec:experiments}, natural language produces rich label formations where $\smat$ induces a geometry of concepts that provides semantic interpretation to embedding structures. Here, we demonstrate that meaningful concepts emerge even in one-hot classification—the traditional focus of neural collapse literature—given minimal deviation from perfect balance.

Specifically, consider class imbalance with ratio $R$: $V/2$ majority classes each have $R>1$ times more samples than the $V/2$ minority classes. It can be analytically shown that $\smatbar$ has singular values exhibiting a three-tier structure \citep{thrampoulidis2022imbalance}:
\begin{align}
\sigma_1&=\ldots=\sigma_{V/2-1}=\sqrt{R} \\&> \sigma_{V/2}=\sqrt{(R+1)/2} \\&> \sigma_{V/2+1}=\ldots=\sigma_{V-1}=1.
\end{align}
Moreover, the left singular vectors matrix $\Ub$ takes a sparse block form:
\[
\Ub = \left[\begin{array}{ccc}
\mathbb{F} & -\sqrt{\frac{1}{V}}\mathbf{1} & \mathbf{0} \\
\mathbf{0} & \sqrt{\frac{1}{V}}\mathbf{1} & \mathbb{F}
\end{array}\right]\in\R^{V\times (V-1)},
\]
where $\mathbb{F}\in\R^{V/2\times(V/2-1)}$ is an orthonormal basis of the subspace orthogonal to $\ones_{V/2}$.\footnote{For concreteness, $\mathbb{F}$ can be constructed using the discrete cosine transform matrix, excluding the constant column: $\mathbb{F}[i,j]=\sqrt{\frac{4}{V}}\cdot\cos\left(\frac{\pi(2i-1)j}{V}\right)$ for $i\in[V/2]$, $j\in[V/2-1]$.}

The structure of $\Ub$ reveals three distinct concept types, corresponding to the three tiers of singular values:

\noindent(1) \textbf{First $V/2-1$ columns} (largest singular values): Non-zero only for majority classes, representing distinctions among majority classes.

\noindent(2) \textbf{Middle column} (singular value $\sqrt{(R+1)/2}$): Opposite-signed entries for majority versus minority classes, encoding the majority-minority dichotomy.

\noindent(3) \textbf{Last $V/2-1$ columns} (smallest singular values): Non-zero only for minority classes, capturing distinctions among minority classes.

This structure demonstrates that the network learns concepts in order of singular value magnitude: first majority class distinctions, then the majority-minority split, and finally minority class differences. Fig.~\ref{fig:confusion_steps} shows an experiment on imbalanced MNIST that confirms this semantic interpretation of concepts (columns of $\Ub$) and validates that class imbalance alone suffices for progressive semantic emergence.

\begin{figure*}[h]
\centering
\begin{tikzpicture}

    \node[rotate=90] at (-0.8, 1) {\textbf{True Class}};

    \foreach \img/\step [count=\i from 0] in {
        cm_step_0.png/Step 0,
        cm_step_13.png/Step 13,
        cm_step_30.png/Step 30,
        cm_step_55.png/Step 55,
        cm_step_65.png/Step 60
    } {
        \ifthenelse{\i=4}{
            \def\imgwidth{2.88cm}
        }{
            \def\imgwidth{2.4cm}
        }

        \begin{scope}[xshift={\i*2.6cm}]
            \node at (1.2cm, 2.8cm) {\small \textbf{\step}};

            \node[anchor=south west, inner sep=0] (image) at (0,0) {\includegraphics[width=\imgwidth]{figs_arxiv_final/\img}};

            \begin{scope}[x={(image.south east)}, y={(image.north west)}]
                \node at (0.125, -0.05) {\tiny maj 1};
                \node at (0.375, -0.05) {\tiny maj 2};
                \node at (0.625, -0.05) {\tiny min 1};
                \node at (0.875, -0.05) {\tiny min 2};

                \ifnum\i=0
                    \node[rotate=90] at (-0.07, 0.875) {\tiny maj 1};
                    \node[rotate=90] at (-0.07, 0.625) {\tiny maj 2};
                    \node[rotate=90] at (-0.07, 0.375) {\tiny min 1};
                    \node[rotate=90] at (-0.07, 0.125) {\tiny min 2};
                \fi
            \end{scope}
        \end{scope}
    }

    \node at (6.5cm, -0.5cm) {\textbf{Predicted Class}};

\end{tikzpicture}
\captionsetup{width=\textwidth}
\caption{Experimental illustration of the fact that important concepts are learned first. Confusion matrix evolution during training of a 3-layer convolutional network on an imbalanced MNIST data with $V=4$ classes, 2 majorities, 2 minorities and imbalance ratio $R=10$. The matrices show five snapshots at different training steps (0, 13, 30, 55, 65) for four classes: two majority classes (Maj1, Maj2) with 100 samples each and two minority classes (Min1, Min2) with 10 samples each. Training progresses from left to right: initially classifying all data as Maj1 (Step 0), then correctly identifying majority classes while misclassifying minority classes (Step 13), gradually improving minority class recognition (Step 30), confusion between minority classes only (Step 55), and finally achieving perfect classification by Step 60. 
For this setting, \citet{seli} show the singular values of $\smatbar$ exhibit a three-tier structure:
$
\sigma_1=\ldots=\sigma_{V/2-1}=\sqrt{R} > \sigma_{V/2}=\sqrt{(R+1)/2} > \sigma_{V/2+1}=\ldots=\sigma_V=1.
$
Additionally, 
The structure of $\Ub$ reveals three distinct types of concepts, corresponding to the three tiers of singular values:
(1) First $V/2-1$ columns (non-zero only for majority classes) represent distinctions among majority classes;
(2) Middle column (singular value $\sqrt{(R+1)/2}$) has opposite-signed entries for majority versus minority classes, encoding the majority-minority dichotomy;
(3) Last $V/2-1$ columns (non-zero only for minority classes) capture distinctions among minority classes.
Thus, the evolution of confusion matrices in the figure perfectly aligns with this hierarchical structure.}
\label{fig:confusion_steps}
\end{figure*}

\subsection{Training transformer on \emph{Simplified Tinystories} dataset} \label{sec:train_tf_detail}
For Simplified TinyStories, we train 6-layer and a 12-layer decoder-only transformers, with 8 attention heads, and 1024 hidden dimensions. Training uses the AdamW optimizer with learning rate 1e-3, $\beta_1 = 0.9$, $\beta_2 = 0.999$, and weight decay 0.001. We use batch size 64 and train until convergence (approximately 1000 epochs).

\subsection{On the hierarchical structure of language} \label{sec:hierarchy} 
As discussed in Sec. \ref{sec:orthant}, language semantics exhibit a hierarchical structure, with more comprehensive semantics involving numerous concepts being inherently more detailed. Fig. \ref{fig:hierarchy} below provides an illustration of this hierarchical structure.
\begin{figure*}[h]
    \centering
    \includegraphics[width=0.8\linewidth]{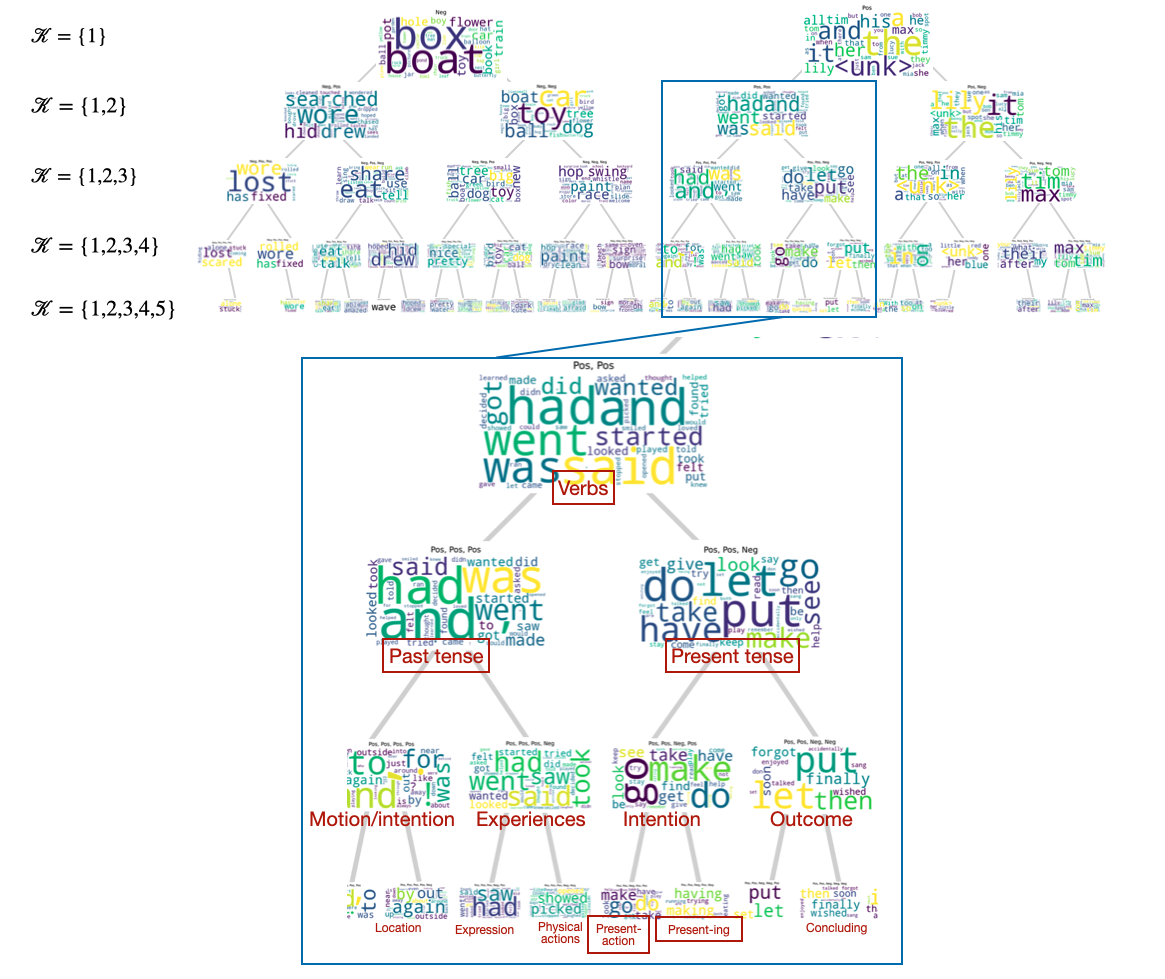}
   \captionsetup{width=\textwidth} \caption{{
   Hierarchical semantic structure revealed by orthant-based clustering on Simplified TinyStories. 
    \textbf{Top:} Clusters are formed by combining 1 to 5 concept dimensions; each row shows word clouds for different sign configurations. Finer-grained semantic categories emerge as more dimensions are added.
    \textbf{Bottom:} Verb-related structure: $\mathcal{K} = {1,2}$ highlights verbs; adding a third dimension separates past and present tense; the fifth dimension further distinguishes verb aspects.}} 
    \label{fig:hierarchy}
\end{figure*}

\subsection{Semantics for Individual Concept}
We include the plots illustrating that individual concept do not contain interpretable semantics as discussed in Sec \ref{sec:experiments}.
\begin{figure*}[h]
    \centering
    \begin{minipage}[t]{0.24\linewidth}
        \centering
        \includegraphics[width=\linewidth]{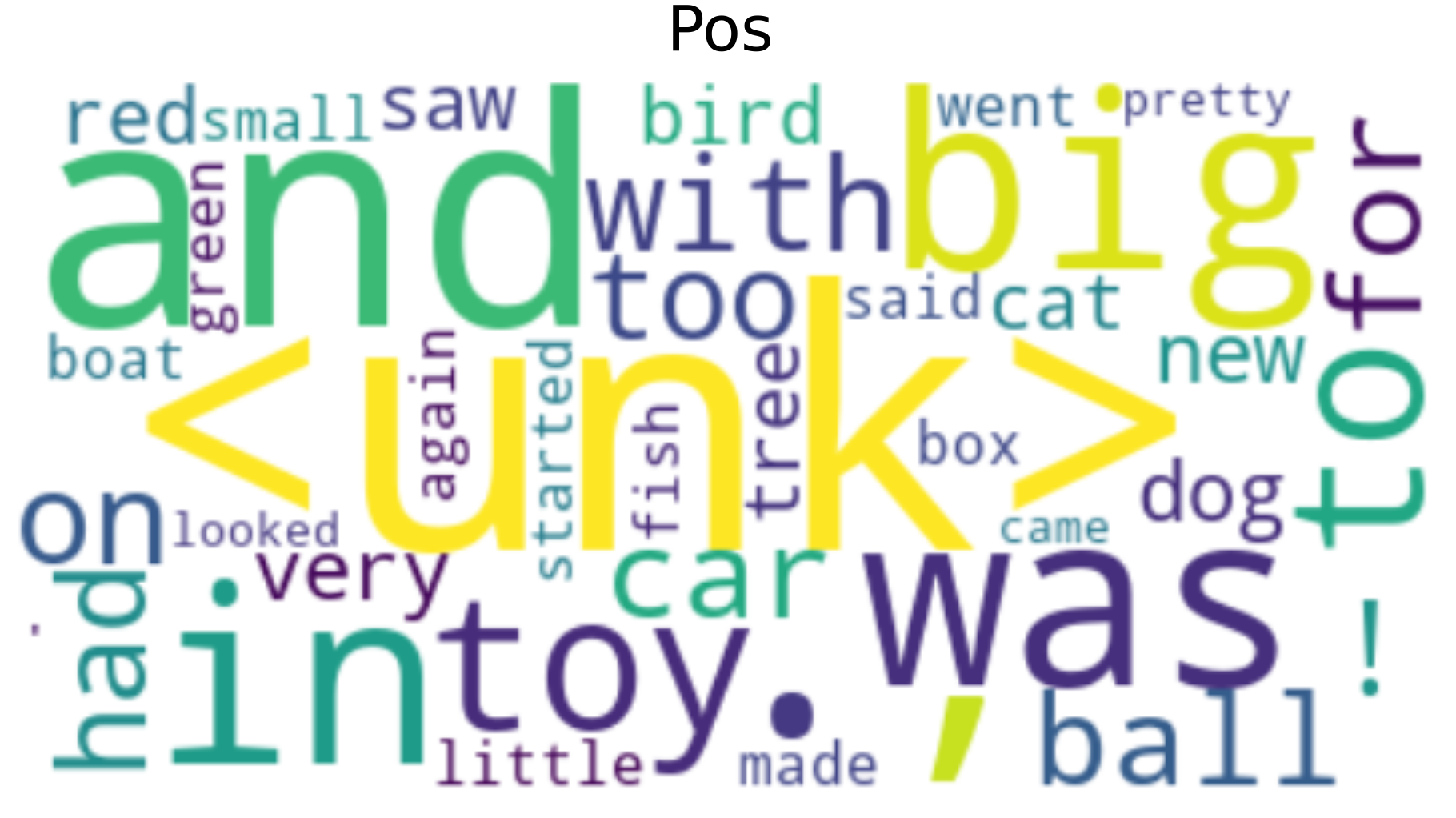}
    \end{minipage}
    \hfill
    \begin{minipage}[t]{0.24\linewidth}
        \centering
        \includegraphics[width=\linewidth]{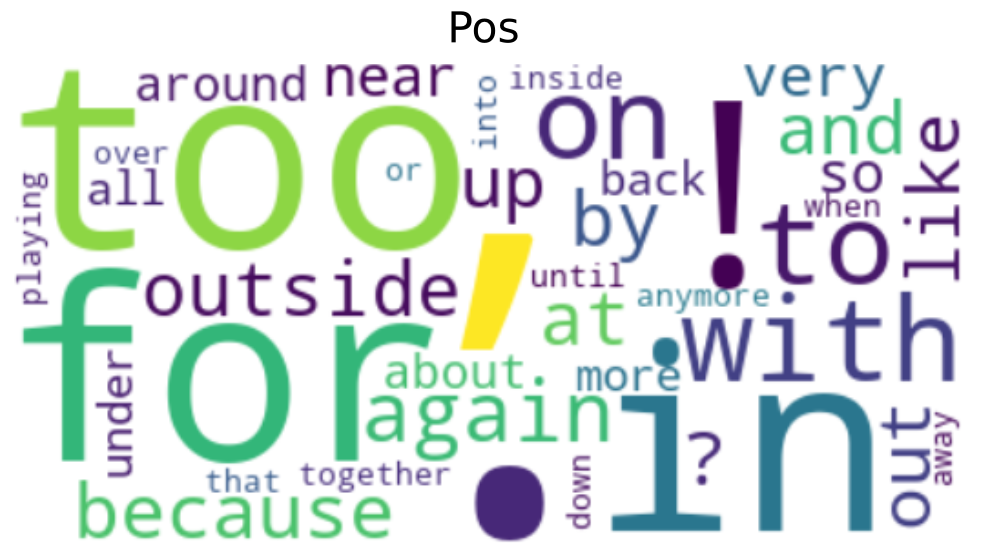}
    \end{minipage}
    \hfill
    \begin{minipage}[t]{0.24\linewidth}
        \centering
        \includegraphics[width=\linewidth]{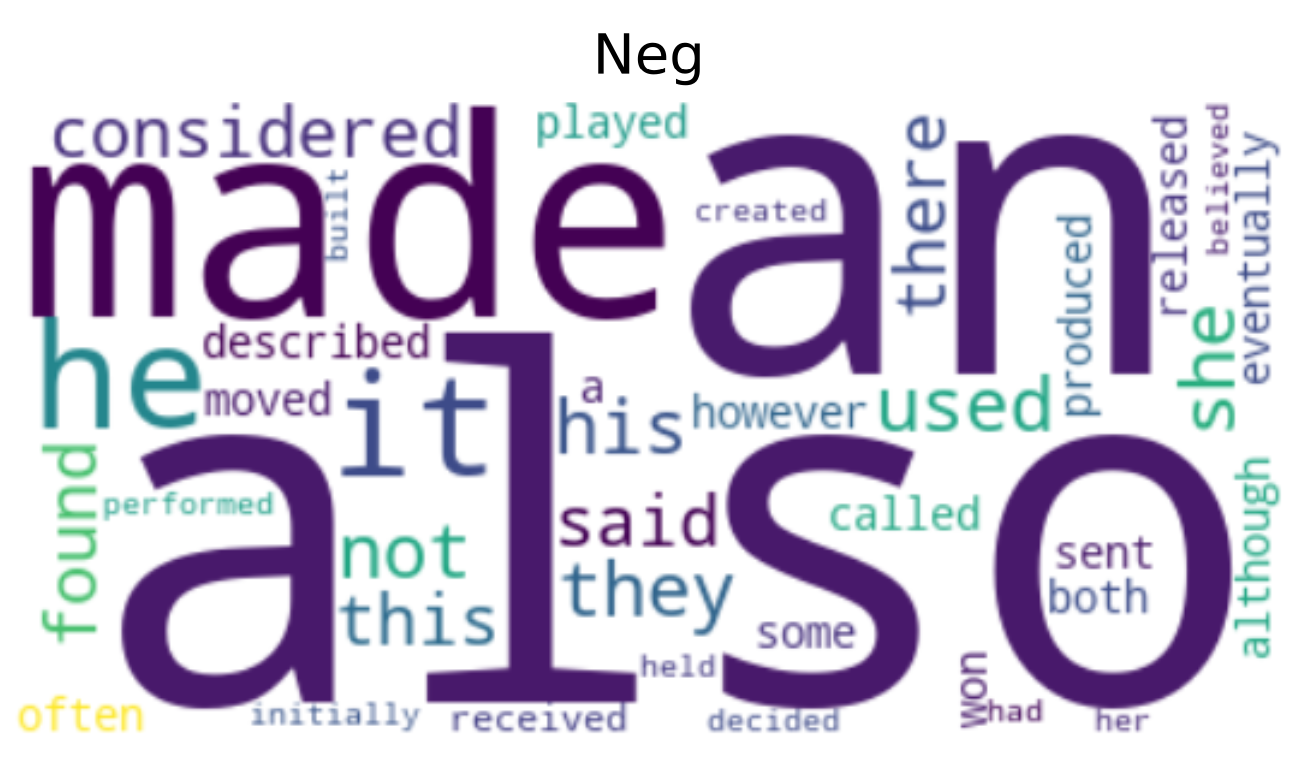}
    \end{minipage}
    \hfill
    \begin{minipage}[t]{0.24\linewidth}
        \centering
        \includegraphics[width=\linewidth]{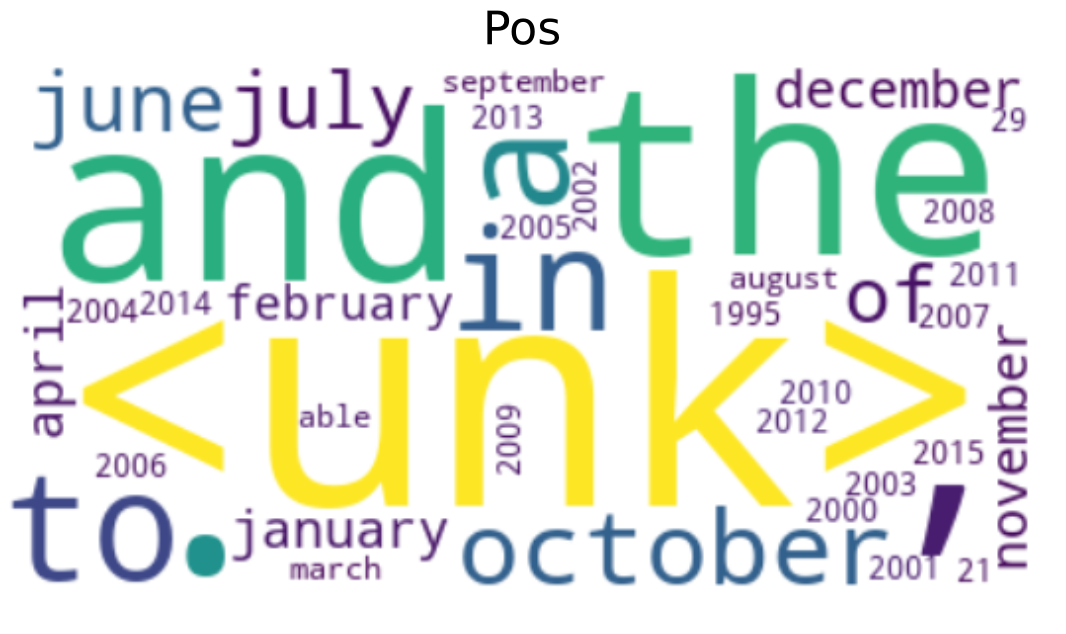}
    \end{minipage}

\captionsetup{width=\textwidth}\caption{Word clouds of top words from individual concept dimensions across datasets, illustrating their lack of human-interpretable semantics without combination. Shown: positive 2nd dimension (Simplified TinyStories), negative 6th dimension (Simplified TinyStories), negative 2nd dimension (Simplified WikiText), positive 4th dimension (Simplified WikiText).
}
    \label{fig:concepts_not_interp}
\end{figure*}



\subsection{Rate of learning}
Fig. \ref{fig:rate} shows the evolution of singular values during training for both squared loss and cross-entropy loss as described in Section \ref{sec:rate}. 
\begin{figure*}[h]
    \centering
    \begin{minipage}{0.48\textwidth}
        \centering
        \begin{tikzpicture}
            \node[anchor=south west,inner sep=0] (image) at (0,0) {
                \includegraphics[width=0.9\textwidth]{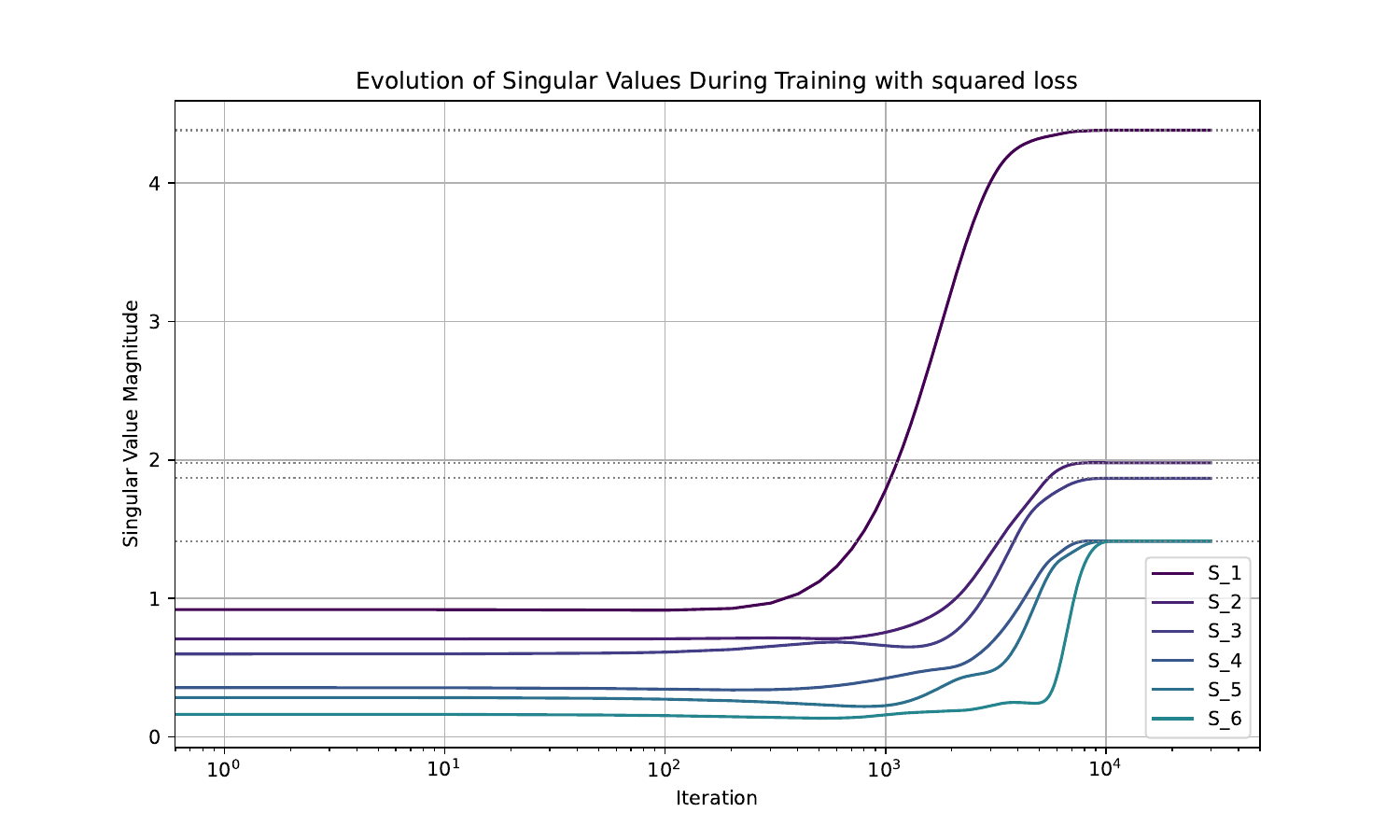} 
            };
            \begin{scope}[x={(image.south east)},y={(image.north west)}]
            \end{scope}
        \end{tikzpicture}
    \end{minipage}\hfill
    \begin{minipage}{0.48\textwidth}
        \centering
        \begin{tikzpicture}
            \node[anchor=south west,inner sep=0] (image) at (0,0) {
                \includegraphics[width=0.9\textwidth]{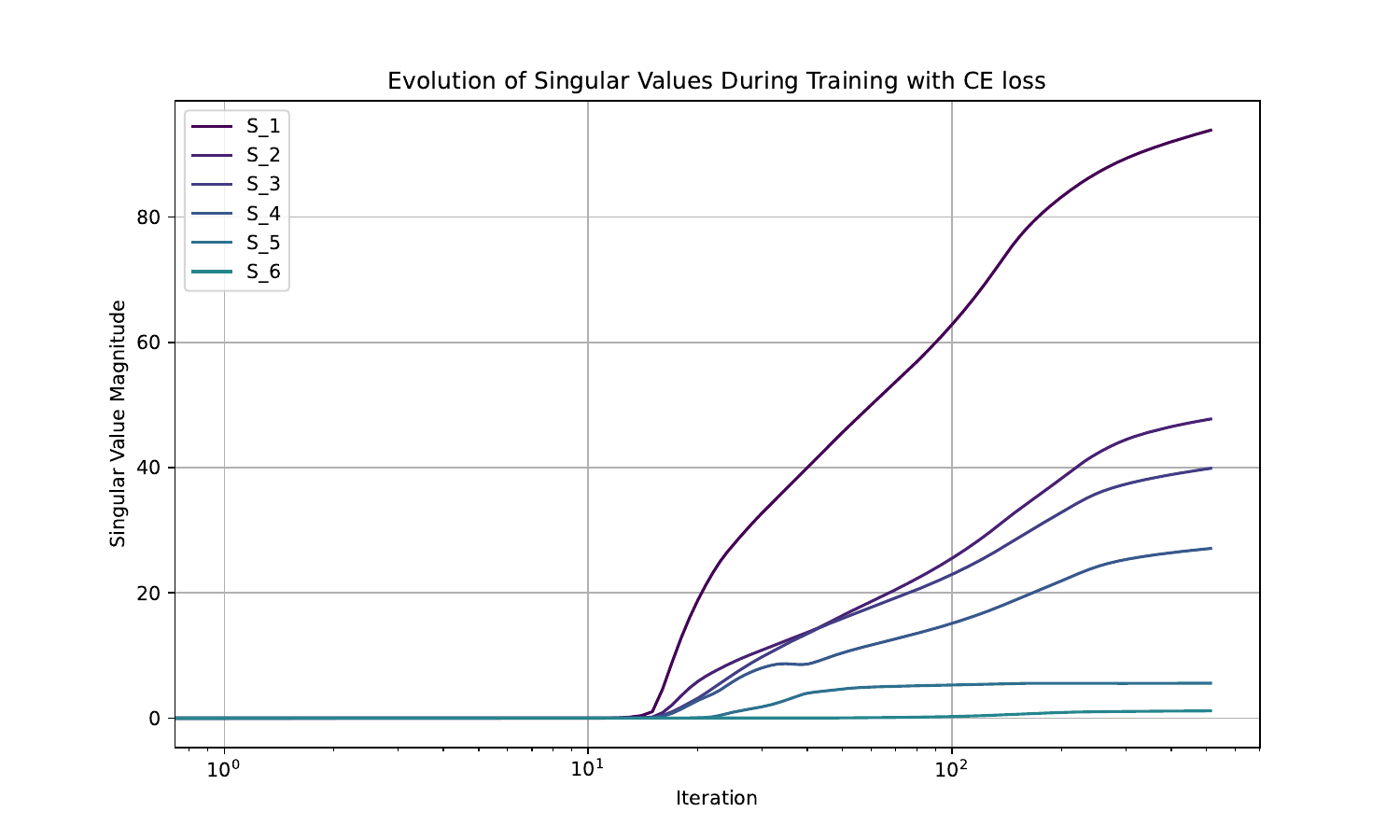} 
            };
            \begin{scope}[x={(image.south east)},y={(image.north west)}]
            \end{scope}
        \end{tikzpicture}
    \end{minipage}
        \captionsetup{width=\textwidth}
\caption{Evolution of singular values of the logit matrix during training. Both plots show that dominant concepts (corresponding to larger singular values) are learned first. \textbf{(Left)} With squared loss, singular values converge to those of the optimal solution, demonstrating learning saturation. \textbf{(Right)} With CE loss, singular values grow unboundedly while maintaining their relative ordering, reflecting the continuous growth of embedding norms characteristic of CE training.}
\label{fig:rate}
\end{figure*}

\subsection{Extension to Masked Language Models} \label{sec:bert}


To explore semantic structure in model trained with different objectives, we applied orthant-based clustering to token embeddings from a pretrained BERT model. As shown in Fig.~\ref{fig:bert_orthant}, the method recovers a range of interpretable semantic categories.


\begin{figure*}[h]
    \centering
    \begin{minipage}[t]{0.23\linewidth}
        \centering
        \textbf{\small 3-digit Number} \\
        \includegraphics[width=\linewidth]{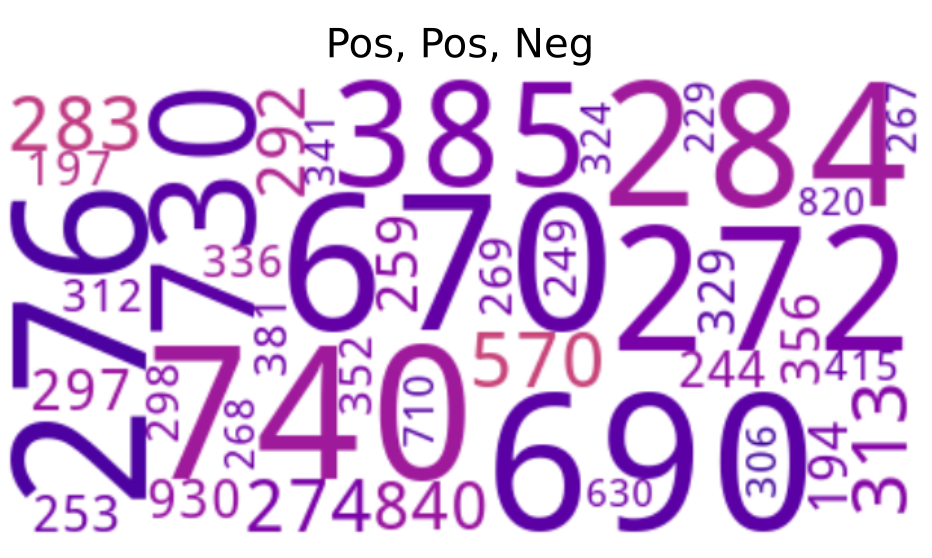}
    \end{minipage}
    \hfill
    \begin{minipage}[t]{0.23\linewidth}
        \centering
        \textbf{\small Historical Years} \\
        \includegraphics[width=\linewidth]{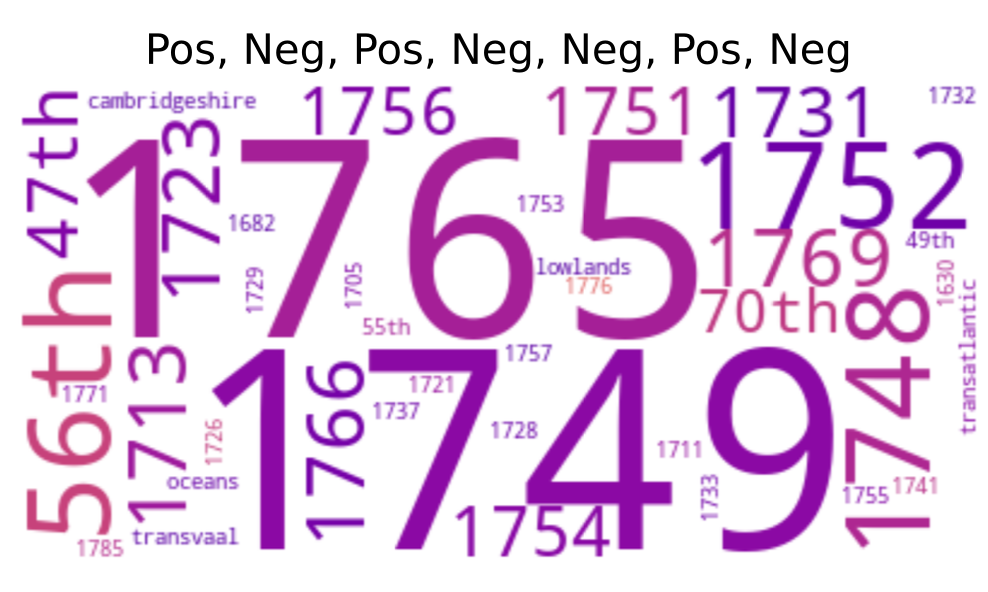}
    \end{minipage}
    \hfill
    \hfill
    \begin{minipage}[t]{0.23\linewidth}
        \centering
        \textbf{\small Action} \\
        \includegraphics[width=\linewidth]{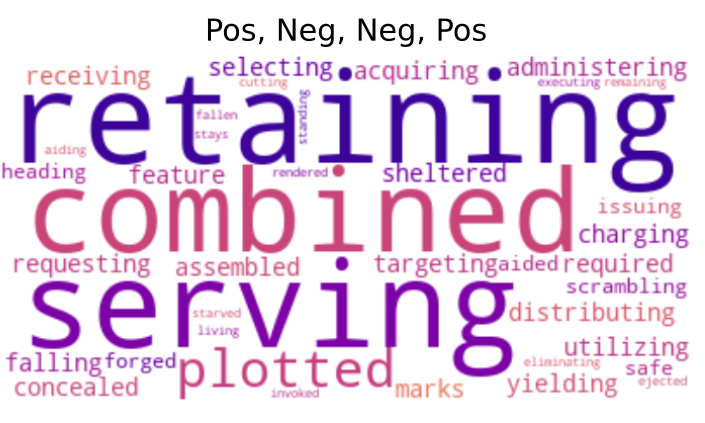}
    \end{minipage}
    \hfill
    \begin{minipage}[t]{0.23\linewidth}
        \centering
        \textbf{\small Administration} \\
        \includegraphics[width=\linewidth]{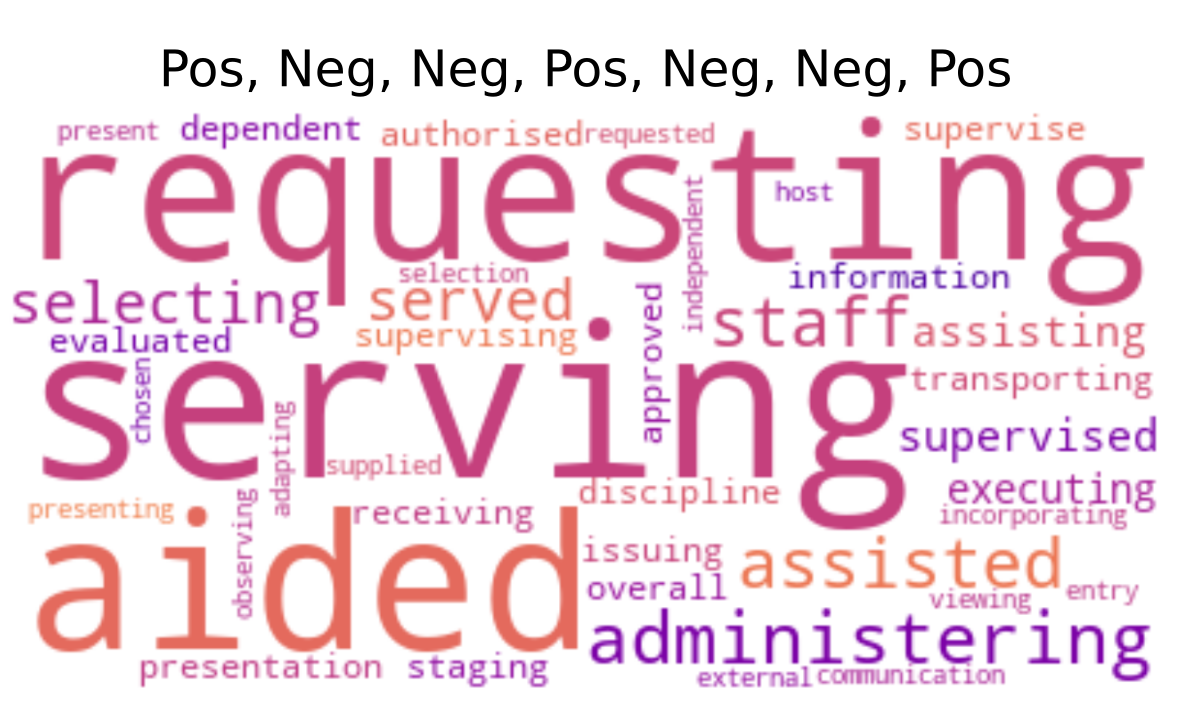}
    \end{minipage}


    \begin{minipage}[t]{0.23\linewidth}
        \centering
        \textbf{\small Given Names} \\
        \includegraphics[width=\linewidth]{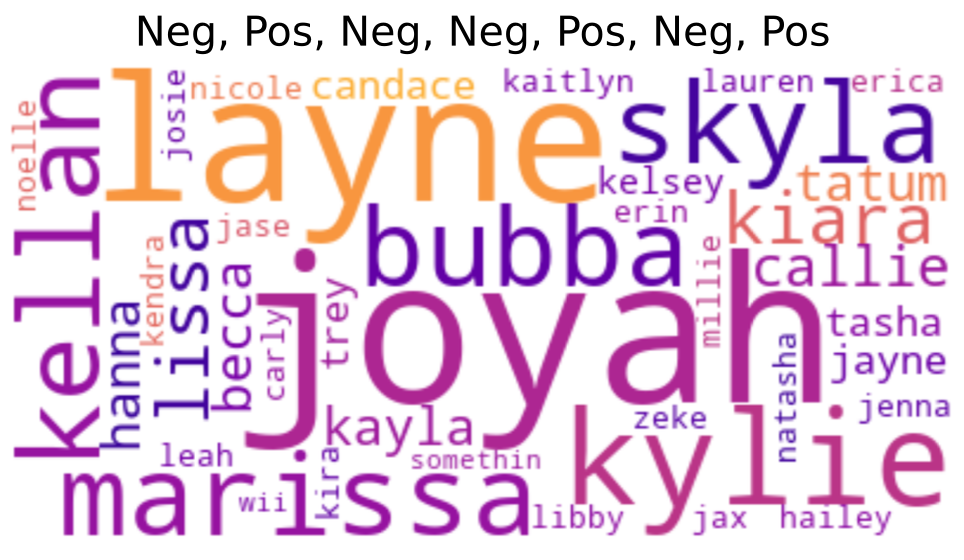}
    \end{minipage}
    \hfill
    \begin{minipage}[t]{0.23\linewidth}
        \centering
        \textbf{\small Operations} \\
        \includegraphics[width=\linewidth]{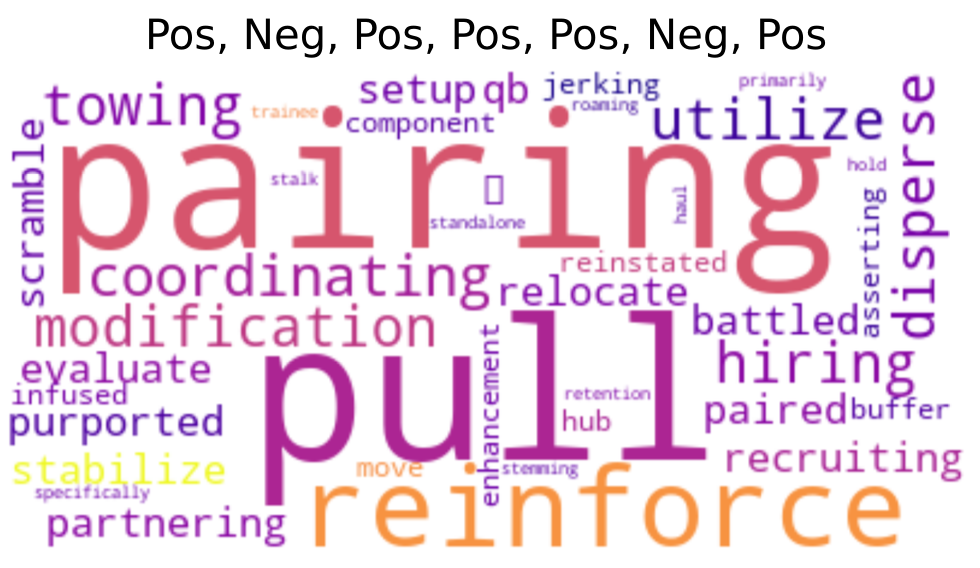}
    \end{minipage}
    \hfill
    \hfill
    \begin{minipage}[t]{0.23\linewidth}
        \centering
        \textbf{\small Name Suffixes} \\
        \includegraphics[width=\linewidth]{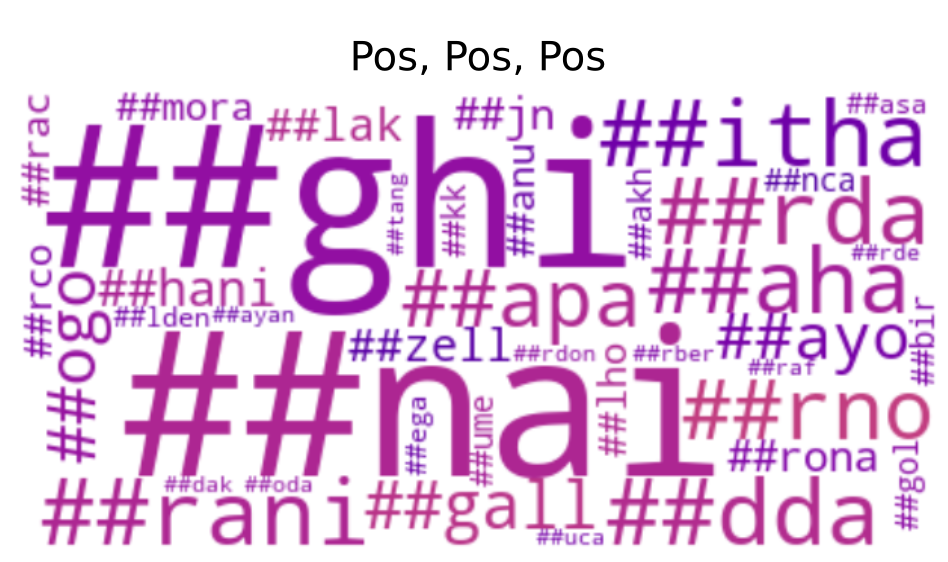}
    \end{minipage}
    \hfill
    \begin{minipage}[t]{0.23\linewidth}
        \centering
        \textbf{\small Verb Suffixes} \\
        \includegraphics[width=\linewidth]{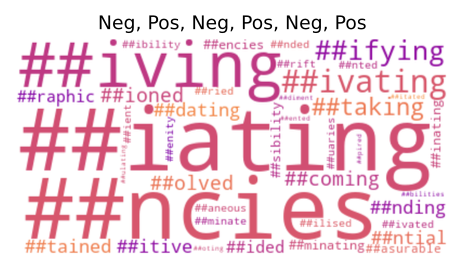}
    \end{minipage}
    \captionsetup{width=\textwidth}
    \caption{ Semantics identified by orthant-based clustering on BERT's word embeddings. In BERT's tokenizer, tokens prefixed with \protect\texttt{\#\#} represent subword units that appear within or at the end of a word.
 }
    \label{fig:bert_orthant}
\end{figure*}

\subsection{Extension to Higher-Dimensional Embeddings} \label{sec:qwen}

As discussed in Sec.~\ref{sec:exp_findings},  we extend our analysis to the Qwen-7B model. We summarize representative semantic categories for selected orthant configurations in Table~\ref{tab:qwen}.  


\begin{table*}[ht]
    \centering
    \includegraphics[width=0.9\linewidth]{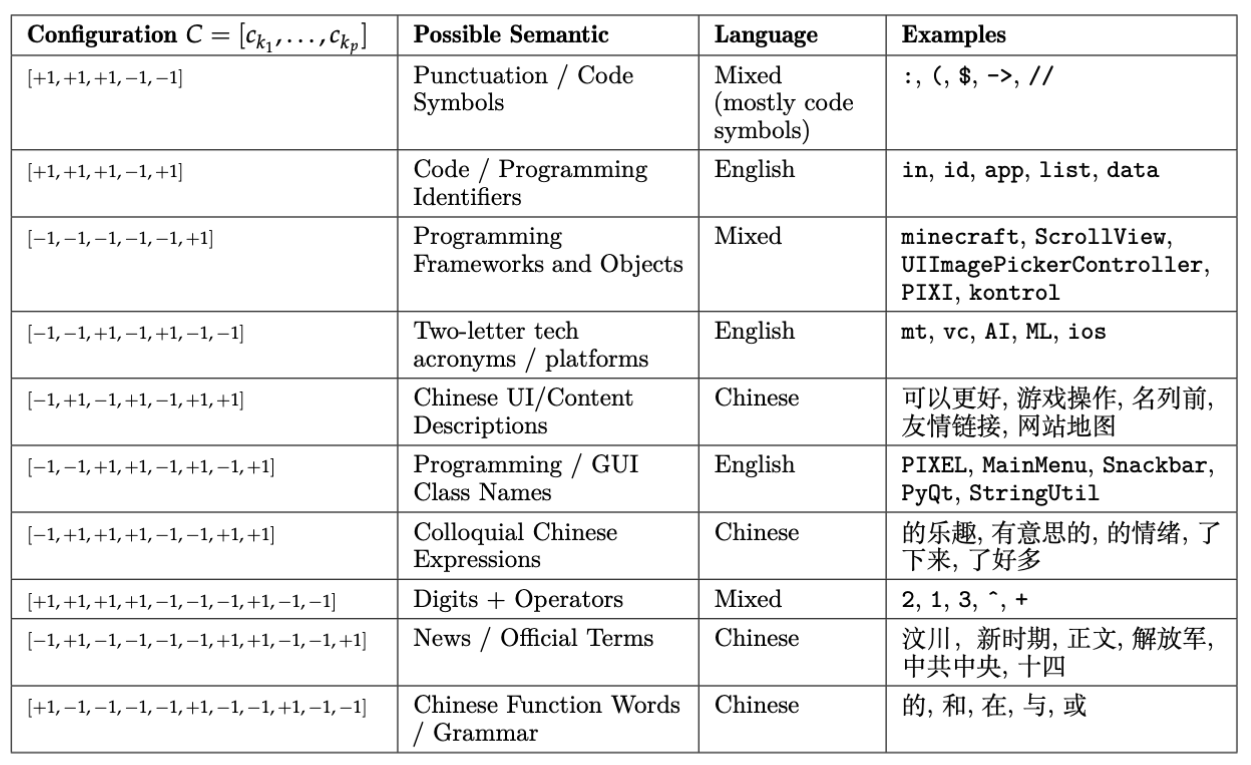}
    \caption{Examples of semantic categories recovered from Qwen-7B using orthant-based clustering on its 4096-dimensional word embeddings. Each row corresponds to a specific sign configuration \( C = [c_{k_1}, \ldots, c_{k_p}] \) along the top singular concept directions. }
    \label{tab:qwen}
\end{table*}


\subsection{Semantics from K-means Spectral Clustering on Simplified TinyStories and Wiki-Text} \label{sec:app-kmeans}
Unlike orthant-based clustering, which can produce up to $2^p$ clusters when selecting $p$ top concepts, vanilla k-means on principal components results in $p$ clusters. Thus, when evaluating spectral methods, we choose $k$ on the order of $2^p$ for fair comparison.
Specifically We apply $k$-means clustering using pairwise distances of analyzer vectors restricted to their top $p = \log_2(k)$ dimensions. As mentioned, this choice corresponds to the top-$p$ concept directions anddiffers from vanilla spectral clustering, which selects $k$ principal components. We chose $k=32$ for Simplified TinyStories and $k=64$ for WikiText-2. Fig.~\ref{fig:kmeans} displays example clusters. 

\begin{figure*}[h]
    \centering
    \begin{minipage}[t]{0.23\linewidth}
        \centering
        \textbf{\small Past-tense verbs} 
        \includegraphics[width=\linewidth]{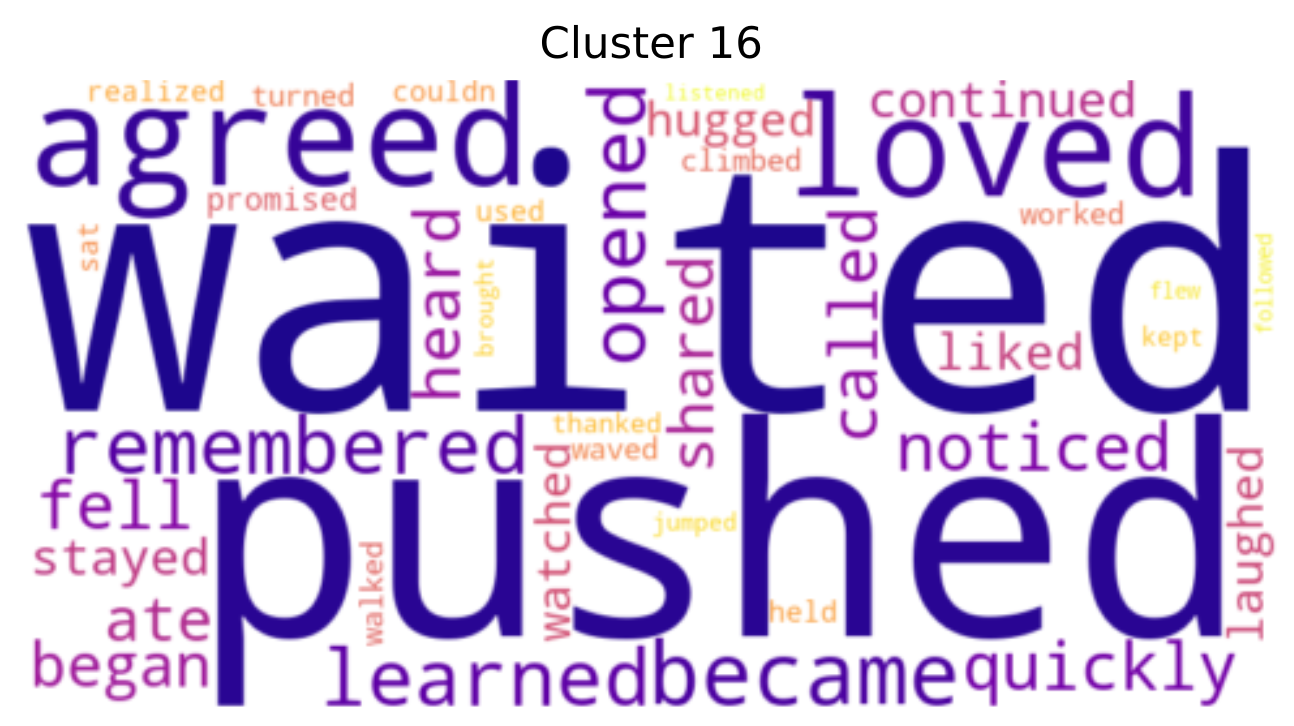}
    \end{minipage}
    \hfill
    \begin{minipage}[t]{0.23\linewidth}
        \centering
        \textbf{\small Proper names} 
        \includegraphics[width=\linewidth]{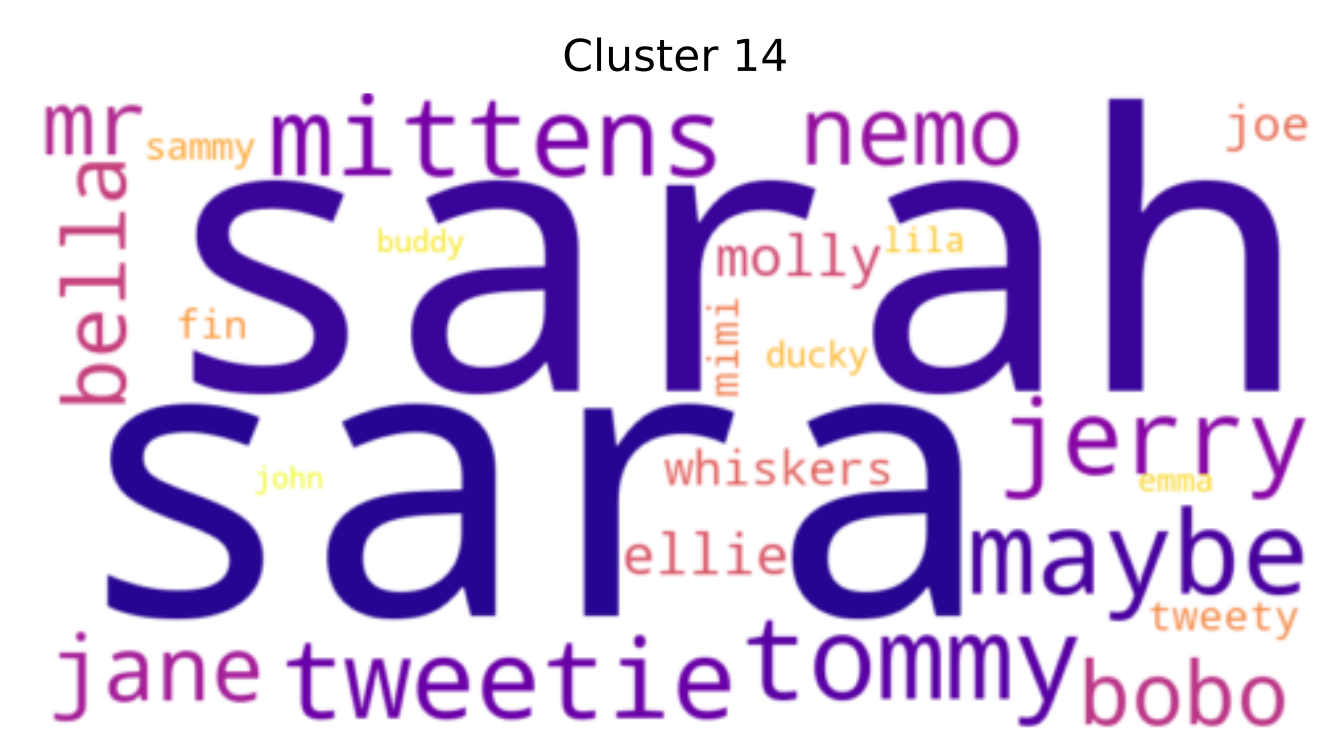}
    \end{minipage}
    \hfill
\begin{minipage}[t]{0.01\linewidth}  
    \centering
    \vspace{1.2em}  
    \rule{0.5pt}{1.5cm}  
\end{minipage}  
    \hfill
    \begin{minipage}[t]{0.22\linewidth}
        \centering
        \textbf{\small Numbers} 
        \includegraphics[width=\linewidth]{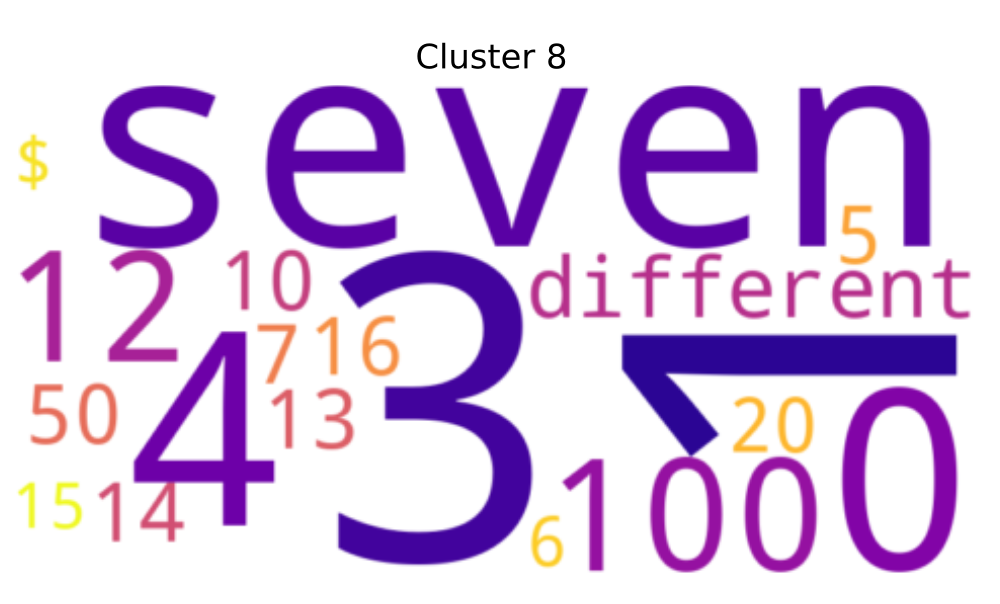}
    \end{minipage}
    \hfill
    \begin{minipage}[t]{0.23\linewidth}
        \centering
        \textbf{\small Conjunctions} 
        \includegraphics[width=\linewidth]{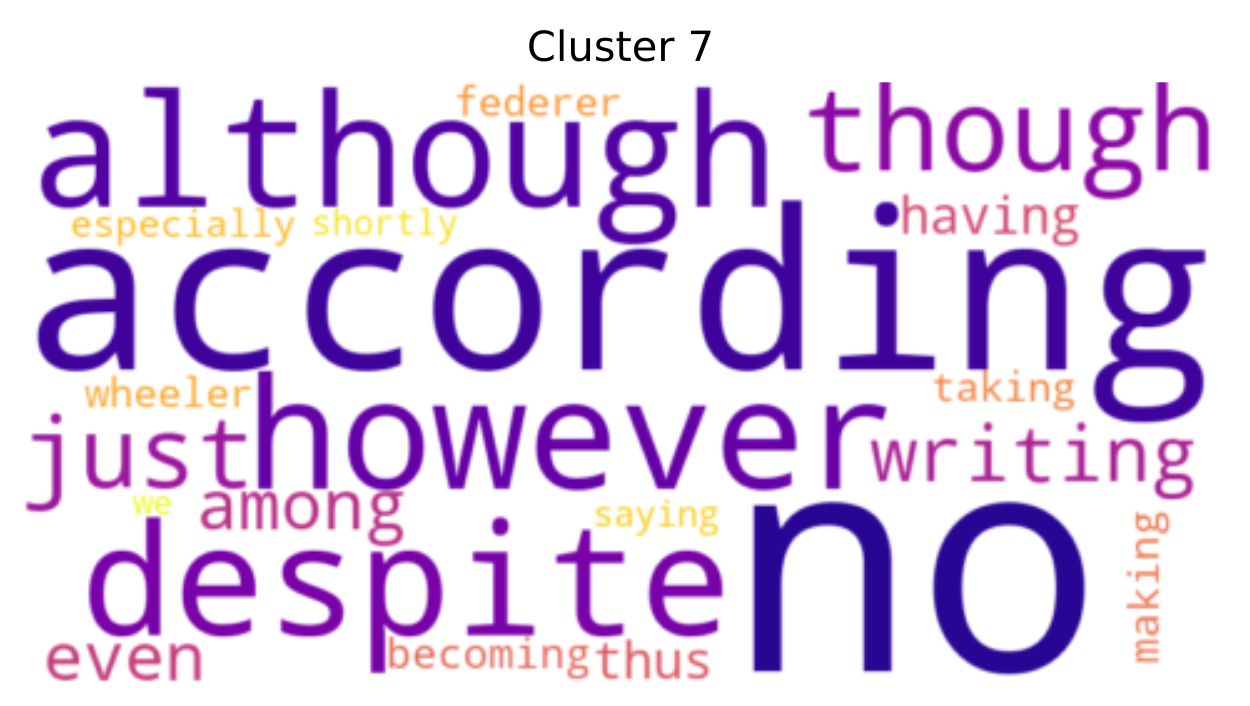}
    \end{minipage}

    \captionsetup{width=\textwidth}
    \caption{$k$-means cluster with top-$p=\log_2(k)$ dimensions of word-analyzer vectors. \emph{Left:} 2 of 32-means clusters with $\Kc=\{1,\ldots,5\}$ on Simplified TinyStories representing Past-tense verbs and Proper names; \emph{Right:} 2 of 64-means cluster with $\Kc=\{1,\ldots,6\}$ on simplified WikiText, representing Numbers and Conjunctions. }

    \label{fig:kmeans}
    \end{figure*}

\section{Use of AI Assistants}
We used AI assistants (ChatGPT, Claude, and GitHub Copilot) to support manuscript preparation. ChatGPT and Claude assisted with text editing, grammar correction, and phrasing refinement. GitHub Copilot assisted with code generation and debugging. All conceptual contributions, theoretical results, experimental designs, and scientific conclusions are solely the work of the authors. AI-generated content was critically reviewed by the authors.

\end{document}